\documentclass{article}

\usepackage[preprint]{neurips_2026}
\usepackage[utf8]{inputenc}
\usepackage[T1]{fontenc}
\usepackage{url}
\usepackage{booktabs}
\usepackage{amsfonts}
\usepackage{microtype}
\usepackage{xcolor}
\usepackage{graphicx}
\usepackage{amsmath}
\usepackage{amssymb}
\usepackage{mathtools}
\usepackage{amsthm}
\usepackage{algorithm}
\usepackage{algorithmic}
\usepackage{tabularx}

\theoremstyle{plain}

\theoremstyle{definition}

\theoremstyle{remark}

\newcommand{\Std}[1]{{\scriptsize \textcolor{gray}{$\pm$ #1}}}

\renewenvironment{abstract}%
{%
  \vskip 0.075in%
  \centerline{\large\bf Abstract}%
  \vspace{0.5ex}%
  \begin{list}{}{%
    \setlength{\leftmargin}{2.5pc}%
    \setlength{\rightmargin}{2.5pc}%
  }\item\relax
}%
{%
  \par\end{list}%
  \vskip 1ex%
}

\title{FlexLAM: Resolving the Bottleneck Trade-off in Latent Action Learning}

\author{%
  Takanori Yoshimoto$^{1\,*}$, \quad
  Yang Hu$^{2}$, \quad
  Naruya Kondo$^{1}$, \quad
  Tatsuya Matsushima$^{2\,*}$ \\
  $^1$University of Tsukuba, \quad $^2$The University of Tokyo \\
  $^*$ Corresponding authors \\
  \url{https://yn35.github.io/flexible-latent-action/}
}

\begin{document}

\maketitle

\begin{abstract}
Latent actions provide a compact interface between action-free video and downstream decision-making, yet existing Latent Action Models (LAMs) force every transition through a fixed-capacity bottleneck. We identify a bottleneck trade-off: overly tight codes can discard transition cues needed for action alignment, while overly loose codes preserve additional transition variation that must be resolved when alignment labels are scarce or narrowly distributed. FlexLAM replaces this fixed capacity with variable-length latent actions trained by nested dropout, yielding prefix-valid codes that capture compact transition structure first and add detail only when needed, without new architectures or losses. A single FlexLAM matches or surpasses separately trained fixed-capacity LAMs at every evaluated token budget under standard scarce-label supervision and under a low-return single-task alignment stress test, indicating that FlexLAM is not merely adjustable at inference time but learns a better latent-action interface at the same token budgets. The same model supports inference-time token-budget adjustment without retraining, and FlexLAM improves Ego4D transition reconstruction. These results suggest that variable-length latent actions are an architecture-free, drop-in upgrade to the fixed-capacity bottleneck in latent action models, latent-action world models, and video-pretrained action interfaces.
\end{abstract}

\section{Introduction}

Latent actions provide a compact interface between action-free video and downstream
decision-making. A Latent Action Model (LAM) compresses an observation transition
$(o_t,o_{t+1})$ into a latent code learned from action-free video, then aligns this
code with executable actions using a smaller labeled set
\citep{edwards2019imitatinglatentpoliciesobservation,rybkin2019learningdoing,menapace2021playablevideogeneration,schmidt2024learningactactions,ye2025latentactionpretrainingvideos,nikulin2025latentactionlearningrequires,chen2025motolatentmotiontoken,chen2025villaxenhancinglatentaction}.
This setting is attractive because action-free videos are abundant, whereas action
labels are costly and often concentrated around particular tasks or embodiments
\citep{embodimentcollaboration2025openxembodimentroboticlearning,black2026pi0visionlanguageactionflowmodel,kim2024openvlaopensourcevisionlanguageactionmodel, bruce2024geniegenerativeinteractiveenvironments}.
Recent in-the-wild latent-action world models study how latent actions can support
world modeling when videos contain richer action variation, environmental noise, and
no common embodiment~\citep{garrido2026learninglatentactionworld}.

\begin{figure}[ht]
  \centering
  \includegraphics[width=0.85\linewidth]{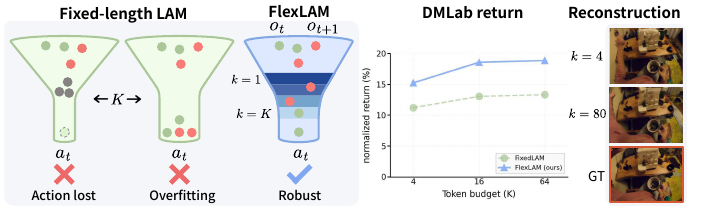}
  \caption{\textbf{The fixed-capacity bottleneck trade-off.}
  \textbf{Left:} One fixed transition-code budget must serve transitions of varying
  complexity, creating tight- and loose-capacity failure modes for action alignment
  under limited labels.
  \textbf{Right:} The main DMLab result: one FlexLAM model beats separately trained
  Fixed-K baselines at every evaluated token budget.}
  \label{fig:lams_bottleneck}
\end{figure}

A central design choice in this interface is its capacity. Existing LAMs typically
instantiate a fixed-capacity interface in which every transition is represented with
the same latent-action budget. This resembles a fixed-rate tokenizer, but transition
complexity is not fixed. Some transitions involve small camera-stable changes, while
others include viewpoint shifts, occlusions, or fine-grained motion. Recent analyses
further suggest that LAM latents may capture nuisance frame differences in addition
to controllable changes, making bottleneck capacity a central design choice rather
than a mere hyperparameter
\citep{zhang2025latentactionmodelsactually,nikulin2025latentactionlearningrequires,liang2025clamcontinuouslatentaction}.
Figure~\ref{fig:lams_bottleneck} summarizes the capacity mismatch and previews the main DMLab result.

This mismatch exposes a bottleneck trade-off. Tight codes impose useful compression
but can remove transition cues needed for latent-to-action alignment. Loose codes
preserve more transition variation, but then the translator must determine which
variation corresponds to executable actions from scarce or narrowly distributed labels.
Thus, the problem is not simply to tune one global capacity, but to learn transition
codes that remain valid across capacities.

Driven by this diagnosis, FlexLAM changes only how the latent-action bottleneck is
trained. Instead of training one fixed code per transition, FlexLAM trains many
retained prefixes of the same code to support transition decoding and action
alignment. This simple rule makes the interface less dependent on suffix-only
transition variation, which stabilizes alignment when labels are scarce or narrowly
distributed. As a consequence, the same model also supports multiple current-step
token budgets at inference time. Because the surrounding LAM pipeline is unchanged,
improvements can be attributed to the latent-action interface rather than to a new
downstream evaluator, policy, or world model.

We evaluate this intervention in the standard LAM pipeline with a fixed latent-token
sequence-model evaluator. The main DMLab comparison is direct: for each budget
$k\in\{4,16,64\}$, we train a separate Fixed-K$k$ model and compare it with the
same FlexLAM model evaluated as FlexLAM@$k$. FlexLAM wins at every evaluated budget.
This shows that retained-prefix training does more than expose intermediate budgets;
it improves the latent-action interface at matched budgets. We use latency, Ego4D
reconstruction, and transition-token visualizations as secondary diagnostics.

We summarize three contributions.
\begin{itemize}
  \item We characterize a fixed-capacity bottleneck trade-off in LAMs, where one
  transition-code budget must serve transitions of varying complexity, creating
  capacity-related risks for latent-to-action alignment under scarce or narrowly
  distributed labels.

  \item We introduce retained-prefix training, which makes every prefix of a
  transition code a valid latent action for both transition decoding and
  latent-to-action alignment. These prefix-valid codes let one model span multiple
  token budgets, and only the bottleneck training changes while the rest of the LAM
  pipeline is unchanged.

  \item We show that a single FlexLAM model outperforms separately trained Fixed-K
  baselines at every evaluated token budget on DMLab under scarce labels. We further
  analyze narrow-source alignment, inference-time token-budget trade-offs, Ego4D
  reconstruction, and cross-embodiment latent-action transfer.
\end{itemize}

\section{Related Work}

\paragraph{Latent action models from action-free video.}
Latent Action Models (LAMs) learn transition codes from action-free observation
pairs and later align these codes with executable actions using a smaller labeled
set~\citep{edwards2019imitatinglatentpoliciesobservation,rybkin2019learningdoing,menapace2021playablevideogeneration,schmidt2024learningactactions,ye2025latentactionpretrainingvideos,nikulin2025latentactionlearningrequires,chen2025motolatentmotiontoken,chen2025villaxenhancinglatentaction}.
Existing LAMs differ in whether their latent actions are discrete or continuous, how
they are aligned to actions, and how they are used downstream. However, most of
them instantiate a fixed-capacity transition-code interface. Recent analyses have
also questioned what LAM codes capture. They may encode controllable transition
structure, but can also reflect nuisance frame differences or distractors
~\citep{zhang2025latentactionmodelsactually,nikulin2025latentactionlearningrequires,liang2025clamcontinuouslatentaction}.
FlexLAM keeps the standard LAM pipeline and studies a complementary representation
question of whether every transition should be forced through the same latent-action
capacity.

\paragraph{Variable-capacity and nested representations.}
Variable-capacity representations have been studied through nested and elastic
embeddings, including nested dropout~\citep{rippel2014learningorderedrepresentationsnested,koikeakino2020stochasticbottleneckratelessautoencoder} and Matryoshka representations
~\citep{kusupati2024matryoshkarepresentationlearning}. Recent tokenization methods similarly adapt the
number of tokens to input complexity, representing simple inputs with fewer tokens
and complex inputs with more~\citep{bachmann2025flextokresamplingimages1d,shen2025catcontentadaptiveimagetokenization}.
Ordered action tokenization has also been explored for autoregressive robot
policies~\citep{liu2026oatorderedactiontokenization}. FlexLAM applies this principle to action-free
transition codes rather than image tokens or policy action tokens. Because transition
complexity varies, the capacity of the latent-action interface should also vary. We use
nested dropout as a local modification to make retained prefixes valid transition
codes and evaluate whether this helps latent-to-action alignment under scarce or
narrowly distributed labels.

\paragraph{Latent-action world models and bottlenecked transitions.}
World models predict future observations or dynamics from learned states or tokens
~\citep{hafner2024masteringdiversedomainsworld,bruce2024geniegenerativeinteractiveenvironments,gao2025adaworldlearningadaptableworld,cui2023a}.
Latent-action world models add an action-like bottleneck to this prediction interface.
Recent in-the-wild latent-action world models study how latent actions can support
world modeling when videos contain richer action variation, environmental noise, and
no common embodiment~\citep{garrido2026learninglatentactionworld}. FlexLAM is
complementary because it studies the capacity of the latent-action interface itself.

\paragraph{Action-label scarcity and VLA policies.}
Many LAMs are motivated by robot and VLA settings, where action labels are costly
and datasets are often concentrated around particular embodiments or tasks
~\citep{embodimentcollaboration2025openxembodimentroboticlearning,black2026pi0visionlanguageactionflowmodel,kim2024openvlaopensourcevisionlanguageactionmodel,khazatsky2025droidlargescaleinthewildrobot}. LAPA-style work
uses latent actions to pretrain VLA policies and validates the resulting policies
through robot manipulation tasks~\citep{ye2025latentactionpretrainingvideos}. FlexLAM addresses an earlier
representation-level bottleneck that precedes such policy learning by asking how much
transition information a latent action code should retain before it is aligned to
executable actions. We therefore evaluate the latent-action interface itself, rather
than proposing a new robot policy architecture or a direct VLA comparison.

\section{The Fixed-Capacity Bottleneck Trade-off}
\label{sec:problem_analysis}

We isolate the capacity mismatch created by fixed-capacity latent actions. Let
$\mathcal{D}_u=\{(o_t,o_{t+1})\}$ denote action-free transitions and
$\mathcal{D}_e=\{(o_t,o_{t+1},a_t)\}$ a smaller action-labeled set used for
alignment. A LAM represents each transition by a length-$K$ latent-action code
$z_t=(z_{t,1},\ldots,z_{t,K})$. The decoder reconstructs $o_{t+1}$ from
$o_t$ and $z_t$, while a translator maps $z_t$ to executable actions using
$\mathcal{D}_e$.

The token length $K$ determines the capacity of the transition-code interface.
More generally, the effective capacity also depends on the token alphabet or
quantizer used by the LAM; we make the exact bottleneck settings explicit in the
experiments and appendix. Here, the key point is that fixed-capacity LAMs assign one
capacity to every transition.

This fixed capacity yields two predictions we test directly. If the bottleneck is too
tight (P1), the code removes cues needed for latent-to-action alignment, so alignment
should degrade as labels grow scarce (Section~\ref{subsec:sample_efficiency}). If it is
too loose (P2), the translator must select action-predictive information from more
transition variation, so a high-capacity code should be fragile under a narrow labeled
source (Section~\ref{subsec:narrow_single_task_alignment}). Thus the issue is not whether
one fixed capacity is universally best, but whether a single global transition-code
budget is a stable interface for alignment. FlexLAM tests this diagnosis by replacing the
fixed interface with retained-prefix codes while holding the downstream alignment and
evaluation interfaces fixed.

\section{FlexLAM}
\label{sec:flexlam}

FlexLAM modifies only the latent-action bottleneck in the standard LAM pipeline.
The surrounding stages---latent-action pretraining, latent-to-action alignment, and
downstream sequence-model evaluation---are kept fixed across bottleneck designs
whenever possible. This makes FlexLAM a controlled intervention on the representation
interface, rather than a new downstream evaluator, policy architecture, or world
model.

\begin{figure}[t]
  \centering
  \includegraphics[width=\linewidth]{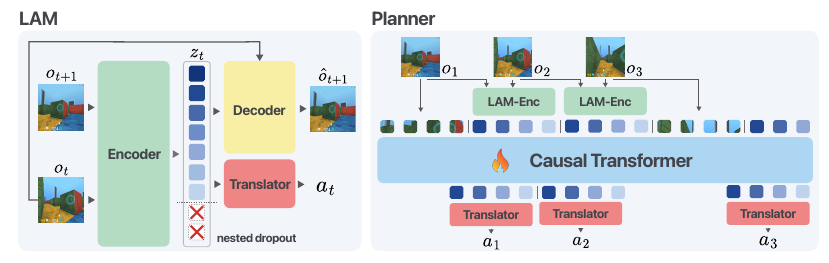}
  \caption{\textbf{FlexLAM overview.}
  (a) During LAM pretraining, FlexLAM samples a retained prefix length $k$ and
  replaces suffix slots with a shared null latent before decoder training.
  (b) The same prefix representation is used for latent-to-action alignment with a
  small labeled set.
  (c) A fixed latent-token evaluator predicts latent-action tokens for
  downstream evaluation using the same translator interface.}
  \label{fig:over_fig}
\end{figure}

\subsection{Retained-Prefix Training for Prefix-Valid Latent Actions}
\label{subsec:prefix_valid_latent_actions}

Let $z_t=(z_{t,1},\ldots,z_{t,K})$ denote the $K$-slot latent-action representation
produced by the LAM encoder and quantizer for a transition $(o_t,o_{t+1})$.
In a fixed-capacity bottleneck, every transition is represented with the same
$K$-slot capacity. FlexLAM instead samples a retained prefix length
$k\sim p(k)$ with $p(k)=\mathrm{Unif}\{0,\ldots,K\}$ and replaces all suffix slots
with a shared learnable null latent $z_{\varnothing}$:
\[
\tilde{z}_{t,j}^{(k)} =
\begin{cases}
z_{t,j}, & j \le k,\\
z_{\varnothing}, & j > k,
\end{cases}
\qquad
\tilde{z}_t^{(k)}\in\mathbb{R}^{K\times d}.
\]
The null latent is optimized with the LAM parameters. The $k=0$ case corresponds to
an all-null training input. This retained-prefix
training follows nested and adaptive representation learning
\citep{rippel2014learningorderedrepresentationsnested,koikeakino2020stochasticbottleneckratelessautoencoder,kusupati2024matryoshkarepresentationlearning,bachmann2025flextokresamplingimages1d}.

We train the LAM encoder and decoder by conditioning the decoder on the current
observation and the null-filled retained prefix:
\[
\min_{\theta,\phi}
\mathbb{E}_{(o_t,o_{t+1})\sim\mathcal{D}_u}
\mathbb{E}_{k\sim p(k)}
\left[
\mathcal{L}_{\mathrm{dec}}(o_{t+1}; o_t,\tilde{z}_t^{(k)})
\right].
\]
Here $\theta$ denotes the LAM encoder-side parameters, including the bottleneck
parameters, and $\phi$ denotes the decoder parameters. The objective
$\mathcal{L}_{\mathrm{dec}}$ is the transition-decoding objective; in our
experiments, it is implemented with a rectified-flow objective in both DMLab and
real-world video settings~\citep{lipman2023flowmatchinggenerativemodeling,liu2022flowstraightfastlearning,esser2024scalingrectifiedflowtransformers}.
In the real-world setting, the decoder is initialized from SD3 and fine-tuned with
the same retained-prefix conditioning principle, enabling higher-resolution
real-video evaluation without changing the core FlexLAM objective.
This objective gives earlier tokens denser training pressure because token
$z_{t,j}$ is retained whenever $k \ge j$. As a result, information useful across
many retained prefixes is encouraged to appear earlier, while later tokens can add
residual transition detail. We use retained-prefix training not only to expose
shorter prefixes at inference time, but also to make the latent-action interface less
sensitive to suffix-only variation during alignment.

\subsection{Latent-to-Action Alignment}
\label{subsec:aligning_la2ea}

To obtain executable actions from latent-action codes learned from action-free video,
we train a translator $g_\psi$ on the labeled set
$\mathcal{D}_e=\{(o_t,o_{t+1},a_t)\}$ to map null-filled retained-prefix
representations to actions. Translator training uses the same prefix sampling
distribution $p(k)$ as LAM pretraining:
\[
\min_{\psi}
\mathbb{E}_{(o_t,o_{t+1},a_t)\sim\mathcal{D}_e}
\mathbb{E}_{k\sim p(k)}
\left[
\ell_{\mathrm{act}}\!\left(g_\psi(\tilde{z}_t^{(k)},a_{t-1}),a_t\right)
\right].
\]
This exposes the translator to many partial views of the same transition code, which
discourages reliance on suffix-only variation when labels are limited.
The translator conditions on the previous action $a_{t-1}$ in all compared methods
to reduce egocentric ambiguity. This is a lightweight action-history cue; related
latent-action systems similarly use proprioceptive cues to ground visually subtle
dynamics~\citep{chen2025villaxenhancinglatentaction}. Different retained prefix
lengths are represented through the null-filled suffix slots, with no separate
length input or attention mask. In our implementation, $g_\psi$ is a fixed-input MLP that
receives the flattened $K$-slot representation. The previous-action conditioning
choice is ablated in Appendix~\ref{app:translator_conditioning}.

The objective above describes the frozen-alignment setting, which isolates the
quality of the latent-action representation by updating only the translator. Prior
work has shown that action supervision during latent-action learning or co-fine-tuning
can improve grounding to executable actions
\citep{nikulin2025latentactionlearningrequires,liang2025clamcontinuouslatentaction,chen2025motolatentmotiontoken}.
We therefore treat joint LAM-translator fine-tuning as a complementary strengthening
of the alignment stage rather than as part of the core FlexLAM intervention. Section~\ref{subsec:joint_training}
shows that this stronger alignment recipe improves bottlenecked models while
preserving FlexLAM's advantage over the fixed-capacity baseline.

\subsection{Latent-Token Sequence Model for Downstream Evaluation}
\label{subsec:latent_token_eval}

We use a fixed causal latent-token evaluator only as a downstream evaluator,
following prior latent-action pipelines
\citep{ye2025latentactionpretrainingvideos,nikulin2025latentactionlearningrequires,chen2025motolatentmotiontoken}. The model predicts
latent-action codes from sparse observation embeddings and past
latent-action blocks. Observation embeddings are used as conditioning inputs rather
than prediction targets.

At decision time, FlexLAM may predict only the first $k$ tokens of the current
latent-action block. The remaining slots are filled with the null latent before the
translator decodes the action, so reducing $k$ shortens only the current-step
autoregressive generation. Full sequence construction, objective, and inference
procedure are provided in Appendix~\ref{app:sequence_model}. Architecture,
quantizer, decoder objectives, and hyperparameter details are reported in
Appendix~\ref{app:dmlab_details} and Appendix~\ref{app:real_world_details}.

\section{Experiments}
\label{sec:results}

We organize the experiments around the bottleneck-interface diagnosis. DMLab is our
controlled downstream setting, where we test whether training many prefixes of the
same transition code to remain useful stabilizes alignment under scarce or narrow
labels.
Ego4D is a complementary real-video reconstruction setting: it tests whether the
same retained-prefix bottleneck yields usable transition representations under real
camera motion and appearance variation. We report the controlled DMLab comparisons
first, then use Ego4D reconstruction as a complementary real-video diagnostic.

\subsection{Experimental Setup}
\label{subsec:experimental_setup}

\paragraph{DMLab.}
We evaluate downstream task performance in DeepMind Lab (DMLab), an egocentric
partially observed environment with viewpoint changes, occlusions, and visual
distractors~\citep{beattie2016deepmindlab}. Expert videos are generated by rolling
out a pretrained DreamerV3 agent~\citep{hafner2024masteringdiversedomainsworld}. The action-free
dataset contains observation transitions $(o_t,o_{t+1})$, and the action-labeled
subset contains $(o_t,o_{t+1},a_t)$. Although simulated, these factors mirror nuisance
variation common in real video, while DMLab still allows controlled downstream return
evaluation.

\paragraph{DMLab baselines and notation.}
We use \textbf{Fixed-K$k$} for a fixed-capacity LAM trained separately with a
$k$-token bottleneck, and \textbf{FlexLAM@$k$} for the same FlexLAM model evaluated
with prefix length $k$. Our main comparison uses $k\in\{4,16,64\}$. All methods use
the same LAM backbone, translator, evaluator architecture, training protocol, and
FSQ vocabulary family; the intended difference is the bottleneck training rule.
This gives a matched-budget test of whether retained-prefix training improves the
latent-action interface. Returns are normalized by the DreamerV3 expert score.

\paragraph{Real-world video.}
For real-world video, we pretrain FlexLAM on a mixture of Internet, egocentric, and
robot videos, including Ego4D, OXE, and other datasets listed in
Appendix~\ref{app:data_mixture_and_sampling}
\citep{grauman2022ego4dworld3000hours,embodimentcollaboration2025openxembodimentroboticlearning}. We compare
against the released villa-X-LAM checkpoint~\citep{chen2025villaxenhancinglatentaction} as an external
fixed-bottleneck LAM reference. villa-X-LAM encodes an 8-frame clip into 7 latent
actions with VQ codebook size 32, and we evaluate it at its native fixed-bottleneck
setting while matching the evaluation fps and input resolution.

\subsection{Sample Efficiency under Scarce Labels}
\label{subsec:sample_efficiency}

We first test prediction~P1 by asking whether retained-prefix training improves
alignment when action labels are extremely scarce. This is the regime where LAMs are most useful because
action-free video can be abundant, but only a small labeled set is available to align
latent codes with executable actions. We pretrain the LAM and latent-token evaluator
on action-free transitions, freeze them, and vary only the amount of labeled data used
to train the translator.

\begin{figure}[t]
  \centering
  \includegraphics[width=\linewidth]{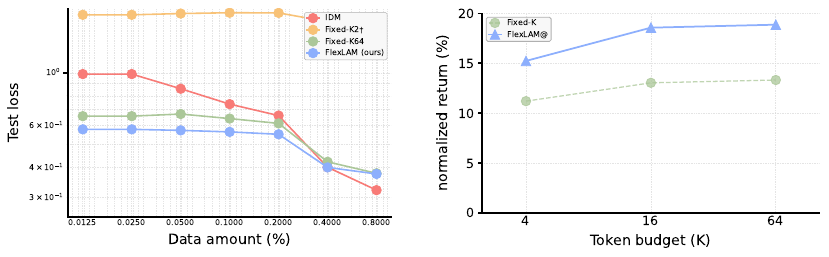}
  \caption{\textbf{Scarce-label alignment and matched-budget return.}
  \emph{Left:} translator test loss versus labeled dataset size.
  \emph{Right:} downstream normalized return under 0.025\% labels at matched token budgets.
  Fixed-$k$ models are trained separately; FlexLAM@$k$ evaluates one FlexLAM model
  at prefix length $k$. FlexLAM outperforms Fixed-K at every evaluated budget.}
  \label{fig:sample_efficiency}
\end{figure}

Figure~\ref{fig:sample_efficiency} provides the full scarce-label comparison previewed in Figure~\ref{fig:lams_bottleneck}.
Across $k\in\{4,16,64\}$, FlexLAM@$k$ outperforms a separately trained Fixed-K$k$ model.
The full-budget result, FlexLAM@64 $>$ Fixed-K64, is especially important: it shows
that the gain is not just access to smaller or intermediate budgets. Retained-prefix
training improves the interface even when the evaluation budget is matched.

\subsection{Action Alignment from a Narrow Single-Task Source}
\label{subsec:narrow_single_task_alignment}

We next evaluate whether FlexLAM remains stable when the labeled alignment set is
drawn from a narrow single-task source. The translator is trained using labels from
Lasertag One Opponent Large, corresponding to 0.04\% of the full dataset. This source
task has low expert return and is excluded from the normalized evaluation suite;
therefore, this setting serves as a practical stress test for action alignment from
a narrow, low-return labeled source.
The source task is excluded from the normalized evaluation suite, so the reported
11-task average evaluates transfer from this narrow source to the remaining tasks.

\begin{figure}[t]
  \centering
  \includegraphics[width=0.93\linewidth]{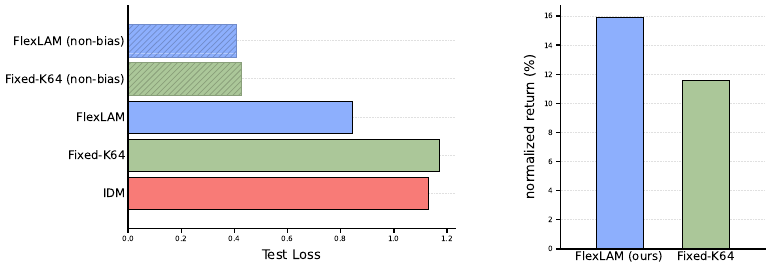}
  \caption{\textbf{Action alignment from a narrow single-task source.}
  The translator is trained using labels from a single low-return source task
  (Lasertag One Opponent Large; 0.04\% of the full dataset), then evaluated on the
  normalized multi-task suite excluding that source task.
  \emph{Left panel} shows translator test loss in this narrow-source setting, compared with a
  control using the same label budget sampled uniformly across tasks.
  \emph{Right panel} shows normalized downstream task return (\% of DreamerV3 expert).
  Full per-task results are reported in
  Appendix~\ref{app:dmlab_normalized_returns}.}
  \label{fig:narrow_single_task_alignment}
\end{figure}

Figure~\ref{fig:narrow_single_task_alignment} confirms prediction~P2. Fixed-capacity
baselines are more fragile under this narrow alignment source. In particular, the high-capacity
Fixed-K64 baseline falls below the random policy on rooms\_watermaze, whereas
FlexLAM avoids this degradation. This is a realistic failure mode for LAMs because
action labels may be sparse and collected from a limited set of behaviors. FlexLAM is
substantially more stable in this setting, outperforming the corresponding
fixed-capacity baselines on most tasks. This suggests that retained-prefix training
makes the interface less dependent on high-capacity suffix slots under narrow labels.

\subsection{Real-World Transition Reconstruction}
\label{subsec:real_world_video_eval}

DMLab evaluates downstream task performance, but the fixed-capacity bottleneck
trade-off is not specific to simulated environments. We therefore evaluate whether
retained-prefix bottlenecks also improve transition reconstruction on visually diverse
real-world video. Ego4D evaluates whether the same bottleneck design improves
real-world transition representation quality, complementing DMLab downstream return
evaluation. We compare FlexLAM against the released villa-X-LAM reference
on Ego4D using per-frame reconstruction metrics, including LPIPS~\citep{zhang2018unreasonableeffectivenessdeepfeatures}.

\begin{table}[t]
\centering
\caption{\textbf{Transition reconstruction on Ego4D.}
Per-frame reconstruction metrics averaged over 200 held-out Ego4D clips.
villa-X-LAM is evaluated using the released checkpoint at its native fixed
bottleneck setting of 7 latent actions per 8-frame clip with VQ codebook size 32.
FlexLAM is evaluated with retained prefix lengths $k\in\{5,20,80\}$ for each
transition.}
\label{tab:ego4d_metrics}
\begin{tabular}{lccc}
\toprule
Method & PSNR$\uparrow$ & SSIM$\uparrow$ & LPIPS$\downarrow$ \\
\midrule
villa-X-LAM         & 16.91 & 0.5442 & 0.5473 \\
FlexLAM ($k{=}5$)   & 18.13 & 0.5758 & 0.3607 \\
FlexLAM ($k{=}20$)  & 19.55 & 0.6358 & 0.3290 \\
FlexLAM ($k{=}80$)  & \textbf{19.84} & \textbf{0.6532} & \textbf{0.3216} \\
\bottomrule
\end{tabular}
\end{table}

Table~\ref{tab:ego4d_metrics} shows that FlexLAM improves over the external
fixed-bottleneck reference across PSNR, SSIM, and LPIPS. Because the two models differ in
pretraining data (ours includes Ego4D), initialization, and nominal capacity, we read
villa-X-LAM as a reference point rather than a controlled baseline; the controlled
evidence here is the within-model trend, where increasing $k$ from 5 to 80 progressively
improves reconstruction quality, consistent with the prefix-valid transition structure
induced by retained-prefix training.

\begin{figure}[t]
  \centering
  \includegraphics[width=\linewidth]{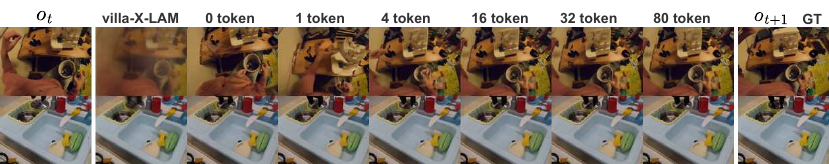}
  \caption{\textbf{Real-world transition reconstruction.}
  We decode latent transition tokens on Ego4D and robot-video reconstruction
  examples. Compared with the released villa-X-LAM reference, FlexLAM produces more
  stable one-step reconstructions under camera and background changes. Varying the
  retained prefix length $k$ within the same model progressively adds visual detail.
  These examples evaluate transition reconstruction under real-video variation.}
  \label{fig:real_world_recon}
\end{figure}

Figure~\ref{fig:real_world_recon} provides representative decoded transitions. The
reference comparison illustrates reconstruction stability under camera and background
changes, while the prefix sweep shows that larger prefixes add visual detail within
the same model. Together with Table~\ref{tab:ego4d_metrics}, these results provide
complementary evidence that the retained-prefix bottleneck also yields usable
transition representations under real-video variation.

\section{Analysis and Ablations}
\label{sec:analysis}

Having shown that FlexLAM improves over separately trained Fixed-K baselines at
matched budgets, we next analyze what the single retained-prefix model provides at
inference time. We first test whether learned latent actions transfer across
embodiments and scenes, then examine how retained prefix length affects translation
loss, downstream return, and latency within the same trained model.

\subsection{Latent Actions Transfer Across Embodiments}
\label{subsec:cross_embodiment_transfer}

\begin{figure}[t]
  \centering
  \includegraphics[width=\linewidth]{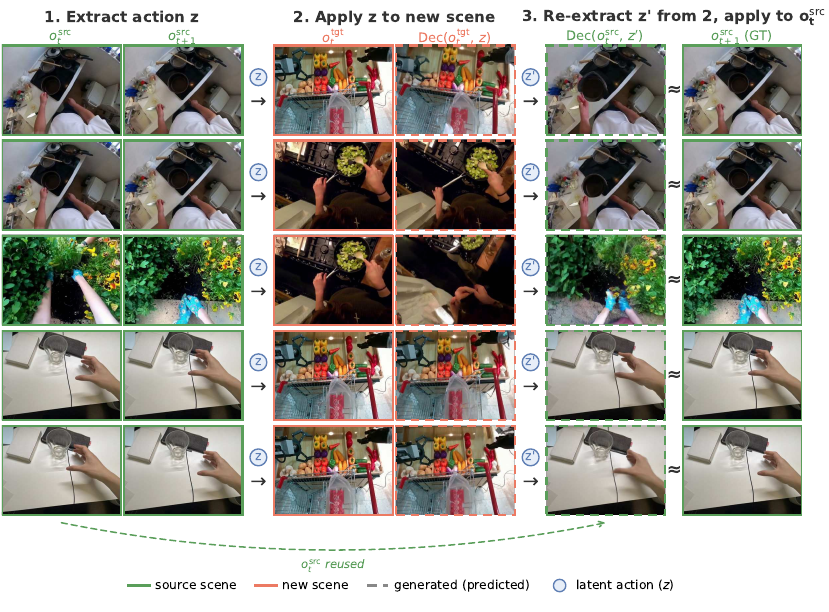}
  \caption{\textbf{Latent actions transfer across embodiments.}
  Each row runs a round trip across two scenes by encoding the source transition,
  $z\,{=}\,\mathrm{Enc}(o_t, o_{t+1})$~(left); decoding it onto the target frame,
  $\hat{o}_{t+1}\,{=}\,\mathrm{Dec}(o_t^{\mathrm{tgt}}, z)$~(middle); re-encoding,
  $z'\,{=}\,\mathrm{Enc}(o_t^{\mathrm{tgt}}, \hat{o}_{t+1})$; and decoding back to the
  source, $\mathrm{Dec}(o_t^{\mathrm{src}}, z')\,{\approx}\,o_{t+1}$~(right).
  Green frames denote the source scene, red the new scene, and dashed frames are
  model-generated.}
  \label{fig:cross_embodiment_transfer}
\end{figure}

We test whether FlexLAM latent actions generalize beyond the source scene.
Figure~\ref{fig:cross_embodiment_transfer} extracts a latent action~$z$ from a
source pair, applies it to a target frame from a different embodiment or scene, and
verifies consistency by re-extracting and applying back to the source.
These pairings span two axes of variation:
morphology (human hands vs.\ robotic grippers) and scene
context (real kitchens and gardens vs.\ tabletop robot setups).
Across all combinations, the target frames consistently reproduce the source
transition---hand movements and object interactions are preserved.
The round-trip recovery closely matches the original~$o_{t+1}$, confirming that
the action information survives the cross-embodiment transfer.
This test is complementary to the reconstruction metrics in
Table~\ref{tab:ego4d_metrics}: reconstruction measures within-domain fidelity,
whereas cross-embodiment transfer tests whether the latent code captures motion
structure independently of visual identity.

\subsection{Retained Prefixes: Alignment and Token-Budget Trade-offs}
\label{subsec:token_budget_tradeoff}

Retained-prefix training exposes multiple operating points within one model.
Reducing $k$ shortens the autoregressive generation for the current
action decision while the historical context is unchanged.
We therefore treat $k$ as a current-step inference budget and measure
the resulting translation, return, and latency trade-offs.

\begin{figure}[t]
  \centering
  \includegraphics[width=0.52\linewidth]{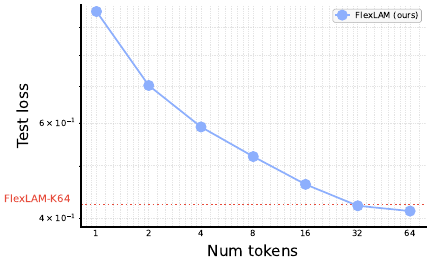}
  \caption{\textbf{Prefix-length scaling within one FlexLAM model.}
  Translator test loss as a function of retained prefix length $k$. This plot varies
  only the retained prefix used by the same trained FlexLAM model. Lower loss
  indicates better latent-to-action alignment.}
  \label{fig:prefix_scaling}
\end{figure}

Figure~\ref{fig:prefix_scaling} shows that intermediate prefixes remain meaningful
operating points within one retained-prefix model. Alignment generally improves as
more tokens are retained, while shorter prefixes remain usable. This analysis is
separate from the matched Fixed-K comparison in Figure~\ref{fig:sample_efficiency};
it characterizes how one FlexLAM model can be operated after training.

\begin{table}[t]
\centering
\caption{\textbf{Current-step token-budget trade-off.}
FlexLAM supports multiple current-step generation budgets within one model.
Latency is measured per decision step under the same inference context and hardware;
full measurement details are provided in Appendix~\ref{app:sequence_model}.}
\label{tab:token_length_analysis}
\begin{tabular}{rccc}
\toprule
$k$ & Norm. Return (\%) & Trans. Loss & Latency (ms/step) \\
\midrule
4  & 20.8 & 0.590 & 57.1 \\
16 & 27.3 & 0.462 & 176.0 \\
64 & 28.8 & 0.413 & 638.8 \\
\bottomrule
\end{tabular}
\end{table}

Table~\ref{tab:token_length_analysis} reports representative operating points of the
same trained FlexLAM model. At $k{=}16$, current-step generation retains 95\% of the
return obtained with full current-step generation while reducing latency by
3.6$\times$. This is a practical consequence of retained-prefix training, whereas the
main evidence for improved representation quality comes from the matched Fixed-K
comparison in Figure~\ref{fig:sample_efficiency}.

\subsection{Joint LAM-Translator Fine-Tuning}
\label{subsec:joint_training}

The main scarce-label experiments use the frozen-alignment setting, which isolates
the quality of the latent-action interface by updating only the translator. This
setting is intentionally controlled, but it is not necessarily the strongest way to
use LAMs when more action labels are available. Because IDM directly observes the
input frames and has no latent bottleneck, it can become competitive with or stronger
than frozen bottlenecked LAM translators as the labeled set grows.

\begin{figure}[t]
  \centering
  \includegraphics[width=0.55\linewidth]{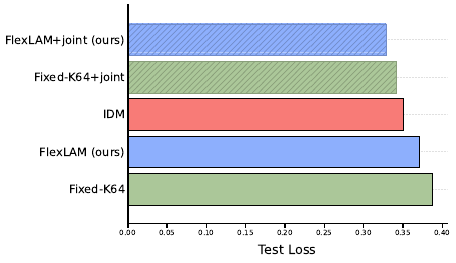}
  \caption{\textbf{Joint LAM-translator fine-tuning.}
  Using 0.5\% action-labeled data, we compare translator validation loss for IDM
  (no bottleneck), a fixed-capacity LAM, and FlexLAM, with and without joint alignment.
  In the frozen setting, IDM can be stronger because it directly observes the input
  frames. Joint alignment allows action loss to update the LAM bottleneck, improving
  bottlenecked LAMs and reversing the IDM-vs-LAM ordering. Under the same joint recipe,
  FlexLAM improves more than the fixed-capacity LAM.}
  \label{fig:joint_training}
\end{figure}

We therefore evaluate the joint-alignment recipe used in prior LAM systems, where
the action loss is allowed to update the LAM encoder and bottleneck parameters
together with the translator. Figure~\ref{fig:joint_training} shows that joint alignment improves bottlenecked
LAMs enough to reverse the frozen-alignment ordering against IDM. This effect is not
specific to FlexLAM: the fixed-capacity LAM also benefits from action-supervised
fine-tuning. The FlexLAM-specific observation is that, under the same joint recipe,
the prefix-valid bottleneck yields a larger improvement than the fixed-capacity
baseline. Thus, retained-prefix training is complementary to joint alignment rather
than a replacement for it.

\section{Discussion and Limitations}

We studied a fixed-capacity bottleneck trade-off in latent action learning, where one
transition-code budget must serve transitions of varying complexity. The main DMLab
finding is that a single retained-prefix FlexLAM model outperforms separately trained
Fixed-K baselines at every evaluated token budget, consistently across both alignment
settings and the eleven-task suite. This suggests that retained-prefix training improves
the latent-action interface itself, rather than only providing additional inference-time
budgets. Ego4D reconstruction and transition-token visualizations provide complementary
representation-level evidence beyond the controlled DMLab setting.

The present evaluation focuses on representation-level evidence: controlled DMLab
alignment and return, Ego4D transition reconstruction, and decoded transition-token
reuse across scenes and embodiments. Two boundaries remain. The controlled downstream
comparison is confined to the DMLab family, and the real-video comparison
(Section~\ref{subsec:real_world_video_eval}) is an external reference rather than a
controlled baseline. Future work should connect these evaluations to real-world
executable policy transfer and action selection under visually ambiguous transitions.

Large-scale video pretraining can involve private, copyrighted, or biased data.
Careful dataset curation, filtering, licensing, and evaluation under distribution shift
remain important before using such representations in downstream systems.

{\small
\bibliographystyle{plainnat}
\bibliography{bibliography}

@inproceedings{kusupati2024matryoshkarepresentationlearning,
 author = {Kusupati, Aditya and Bhatt, Gantavya and Rege, Aniket and Wallingford, Matthew and Sinha, Aditya and Ramanujan, Vivek and Howard-Snyder, William and Chen, Kaifeng and Kakade, Sham and Jain, Prateek and Farhadi, Ali},
 booktitle = {Advances in Neural Information Processing Systems},
 editor = {S. Koyejo and S. Mohamed and A. Agarwal and D. Belgrave and K. Cho and A. Oh},
 pages = {30233--30249},
 publisher = {Curran Associates, Inc.},
 title = {Matryoshka Representation Learning},
 url = {https://proceedings.neurips.cc/paper_files/paper/2022/file/c32319f4868da7613d78af9993100e42-Paper-Conference.pdf},
 volume = {35},
 year = {2022}
}

@inproceedings{
ye2025latentactionpretrainingvideos,
title={Latent Action Pretraining from Videos},
author={Seonghyeon Ye and Joel Jang and Byeongguk Jeon and Se June Joo and Jianwei Yang and Baolin Peng and Ajay Mandlekar and Reuben Tan and Yu-Wei Chao and Bill Yuchen Lin and Lars Liden and Kimin Lee and Jianfeng Gao and Luke Zettlemoyer and Dieter Fox and Minjoon Seo},
booktitle={The Thirteenth International Conference on Learning Representations},
year={2025},
url={https://openreview.net/forum?id=VYOe2eBQeh}
}

@InProceedings{chen2025motolatentmotiontoken, author = {Chen, Yi and Ge, Yuying and Tang, Weiliang and Li, Yizhuo and Ge, Yixiao and Ding, Mingyu and Shan, Ying and Liu, Xihui}, title = {Moto: Latent Motion Token as the Bridging Language for Learning Robot Manipulation from Videos}, booktitle = {Proceedings of the IEEE/CVF International Conference on Computer Vision (ICCV)}, month = {October}, year = {2025}, pages = {19752-19763} }

@inproceedings{
chen2025villaxenhancinglatentaction,
title={villa-X: Enhancing Latent Action Modeling in Vision-Language-Action Models},
author={Xiaoyu Chen and Hangxing Wei and Pushi Zhang and Chuheng Zhang and Kaixin Wang and Yanjiang Guo and Rushuai Yang and Yucen Wang and Xinquan Xiao and Li Zhao and Jianyu Chen and Jiang Bian},
booktitle={The Fourteenth International Conference on Learning Representations},
year={2026},
url={https://openreview.net/forum?id=y5CaJb17Fn}
}

@InProceedings{edwards2019imitatinglatentpoliciesobservation,
  title = 	 {Imitating Latent Policies from Observation},
  author =       {Edwards, Ashley and Sahni, Himanshu and Schroecker, Yannick and Isbell, Charles},
  booktitle = 	 {Proceedings of the 36th International Conference on Machine Learning},
  pages = 	 {1755--1763},
  year = 	 {2019},
  editor = 	 {Chaudhuri, Kamalika and Salakhutdinov, Ruslan},
  volume = 	 {97},
  series = 	 {Proceedings of Machine Learning Research},
  month = 	 {09--15 Jun},
  publisher =    {PMLR},
  url = 	 {https://proceedings.mlr.press/v97/edwards19a.html}
}

@inproceedings{
rybkin2019learningdoing,
title={Learning what you can do before doing anything},
author={Oleh Rybkin and Karl Pertsch and Andrew Jaegle and Konstantinos G. Derpanis and Kostas Daniilidis},
booktitle={International Conference on Learning Representations},
year={2019},
url={https://openreview.net/forum?id=SylPMnR9Ym},
}

@inproceedings{schmidt2024learningactactions,
 author = {Schmidt, Dominik and Jiang, Minqi},
 booktitle = {International Conference on Learning Representations},
 editor = {B. Kim and Y. Yue and S. Chaudhuri and K. Fragkiadaki and M. Khan and Y. Sun},
 pages = {9379--9395},
 title = {Learning to Act without Actions},
 url = {https://proceedings.iclr.cc/paper_files/paper/2024/file/27985d21f0b751b933d675930aa25022-Paper-Conference.pdf},
 volume = {2024},
 year = {2024}
}

@inproceedings{
zhang2025latentactionmodelsactually,
title={What Do Latent Action Models Actually Learn?},
author={Chuheng Zhang and Tim Pearce and Pushi Zhang and Kaixin Wang and Xiaoyu Chen and Wei Shen and Li Zhao and Jiang Bian},
booktitle={The Thirty-ninth Annual Conference on Neural Information Processing Systems},
year={2026},
url={https://openreview.net/forum?id=DQMjemrVhe}
}

@inproceedings{
nikulin2025latentactionlearningrequires,
title={Latent Action Learning Requires Supervision in the Presence of Distractors},
author={Alexander Nikulin and Ilya Zisman and Denis Tarasov and Lyubaykin Nikita and Andrei Polubarov and Igor Kiselev and Vladislav Kurenkov},
booktitle={Forty-second International Conference on Machine Learning},
year={2025},
url={https://openreview.net/forum?id=2gcEQCT7QW}
}

@misc{liang2025clamcontinuouslatentaction,
      title={CLAM: Continuous Latent Action Models for Robot Learning from Unlabeled Demonstrations},
      author={Anthony Liang and Pavel Czempin and Matthew Hong and Yutai Zhou and Erdem Biyik and Stephen Tu},
      year={2025},
      eprint={2505.04999},
      archivePrefix={arXiv},
      primaryClass={cs.RO},
      url={https://arxiv.org/abs/2505.04999},
}

@misc{rippel2014learningorderedrepresentationsnested,
      title={Learning Ordered Representations with Nested Dropout},
      author={Oren Rippel and Michael A. Gelbart and Ryan P. Adams},
      year={2014},
      eprint={1402.0915},
      archivePrefix={arXiv},
      primaryClass={stat.ML},
      url={https://arxiv.org/abs/1402.0915},
}

@inproceedings{koikeakino2020stochasticbottleneckratelessautoencoder, title={Stochastic Bottleneck: Rateless Auto-Encoder for Flexible Dimensionality Reduction}, url={http://dx.doi.org/10.1109/ISIT44484.2020.9174523}, DOI={10.1109/isit44484.2020.9174523}, booktitle={2020 IEEE International Symposium on Information Theory (ISIT)}, publisher={IEEE}, author={Koike-Akino, Toshiaki and Wang, Ye}, year={2020}, month=jun, pages={2735–2740} }

@inproceedings{
kim2024openvlaopensourcevisionlanguageactionmodel,
title={Open{VLA}: An Open-Source Vision-Language-Action Model},
author={Moo Jin Kim and Karl Pertsch and Siddharth Karamcheti and Ted Xiao and Ashwin Balakrishna and Suraj Nair and Rafael Rafailov and Ethan P Foster and Pannag R Sanketi and Quan Vuong and Thomas Kollar and Benjamin Burchfiel and Russ Tedrake and Dorsa Sadigh and Sergey Levine and Percy Liang and Chelsea Finn},
booktitle={8th Annual Conference on Robot Learning},
year={2024},
url={https://openreview.net/forum?id=ZMnD6QZAE6}
}

@InProceedings{peebles2023scalablediffusionmodelstransformers, author = {Peebles, William and Xie, Saining}, title = {Scalable Diffusion Models with Transformers}, booktitle = {Proceedings of the IEEE/CVF International Conference on Computer Vision (ICCV)}, month = {October}, year = {2023}, pages = {4195-4205} }

@misc{chen2025rethinkingshapeconventionmlp,
      title={Rethinking the shape convention of an MLP},
      author={Meng-Hsi Chen and Yu-Ang Lee and Feng-Ting Liao and Da-shan Shiu},
      year={2025},
      eprint={2510.01796},
      archivePrefix={arXiv},
      primaryClass={cs.LG},
      url={https://arxiv.org/abs/2510.01796},
}

@InProceedings{bruce2024geniegenerativeinteractiveenvironments,
  title = 	 {Genie: Generative Interactive Environments},
  author =       {Bruce, Jake and Dennis, Michael D and Edwards, Ashley and Parker-Holder, Jack and Shi, Yuge and Hughes, Edward and Lai, Matthew and Mavalankar, Aditi and Steigerwald, Richie and Apps, Chris and Aytar, Yusuf and Bechtle, Sarah Maria Elisabeth and Behbahani, Feryal and Chan, Stephanie C.Y. and Heess, Nicolas and Gonzalez, Lucy and Osindero, Simon and Ozair, Sherjil and Reed, Scott and Zhang, Jingwei and Zolna, Konrad and Clune, Jeff and Freitas, Nando De and Singh, Satinder and Rockt\"{a}schel, Tim},
  booktitle = 	 {Proceedings of the 41st International Conference on Machine Learning},
  pages = 	 {4603--4623},
  year = 	 {2024},
  editor = 	 {Salakhutdinov, Ruslan and Kolter, Zico and Heller, Katherine and Weller, Adrian and Oliver, Nuria and Scarlett, Jonathan and Berkenkamp, Felix},
  volume = 	 {235},
  series = 	 {Proceedings of Machine Learning Research},
  month = 	 {21--27 Jul},
  publisher =    {PMLR},
  url = 	 {https://proceedings.mlr.press/v235/bruce24a.html}
}

@InProceedings{menapace2021playablevideogeneration, author = {Menapace, Willi and Lathuiliere, Stephane and Tulyakov, Sergey and Siarohin, Aliaksandr and Ricci, Elisa}, title = {Playable Video Generation}, booktitle = {Proceedings of the IEEE/CVF Conference on Computer Vision and Pattern Recognition (CVPR)}, month = {June}, year = {2021}, pages = {10061-10070} }

@inproceedings{
cui2023a,
title={A Universal World Model Learned from Large Scale and Diverse Videos},
author={Hanchen Cui and Yang Gao},
booktitle={NeurIPS 2023 Foundation Models for Decision Making Workshop},
year={2023},
url={https://openreview.net/forum?id=lw5GlytIY5}
}

@InProceedings{gao2025adaworldlearningadaptableworld,
  title = 	 {{A}da{W}orld: Learning Adaptable World Models with Latent Actions},
  author =       {Gao, Shenyuan and Zhou, Siyuan and Du, Yilun and Zhang, Jun and Gan, Chuang},
  booktitle = 	 {Proceedings of the 42nd International Conference on Machine Learning},
  pages = 	 {18744--18771},
  year = 	 {2025},
  editor = 	 {Singh, Aarti and Fazel, Maryam and Hsu, Daniel and Lacoste-Julien, Simon and Berkenkamp, Felix and Maharaj, Tegan and Wagstaff, Kiri and Zhu, Jerry},
  volume = 	 {267},
  series = 	 {Proceedings of Machine Learning Research},
  month = 	 {13--19 Jul},
  publisher =    {PMLR},
  url = 	 {https://proceedings.mlr.press/v267/gao25u.html}
}

@InProceedings{bachmann2025flextokresamplingimages1d,
  title = 	 {{F}lex{T}ok: Resampling Images into 1{D} Token Sequences of Flexible Length},
  author =       {Bachmann, Roman and Allardice, Jesse and Mizrahi, David and Fini, Enrico and Kar, O\u{g}uzhan Fatih and Amirloo, Elmira and El-Nouby, Alaaeldin and Zamir, Amir and Dehghan, Afshin},
  booktitle = 	 {Proceedings of the 42nd International Conference on Machine Learning},
  pages = 	 {2241--2292},
  year = 	 {2025},
  editor = 	 {Singh, Aarti and Fazel, Maryam and Hsu, Daniel and Lacoste-Julien, Simon and Berkenkamp, Felix and Maharaj, Tegan and Wagstaff, Kiri and Zhu, Jerry},
  volume = 	 {267},
  series = 	 {Proceedings of Machine Learning Research},
  month = 	 {13--19 Jul},
  publisher =    {PMLR},
  url = 	 {https://proceedings.mlr.press/v267/bachmann25a.html}
}

@misc{hafner2024masteringdiversedomainsworld,
      title={Mastering Diverse Domains through World Models},
      author={Danijar Hafner and Jurgis Pasukonis and Jimmy Ba and Timothy Lillicrap},
      year={2024},
      eprint={2301.04104},
      archivePrefix={arXiv},
      primaryClass={cs.AI},
      url={https://arxiv.org/abs/2301.04104},
}

@misc{garrido2026learninglatentactionworld,
      title={Learning Latent Action World Models In The Wild},
      author={Quentin Garrido and Tushar Nagarajan and Basile Terver and Nicolas Ballas and Yann LeCun and Michael Rabbat},
      year={2026},
      eprint={2601.05230},
      archivePrefix={arXiv},
      primaryClass={cs.AI},
      url={https://arxiv.org/abs/2601.05230},
}

@InProceedings{grauman2022ego4dworld3000hours, author = {Grauman, Kristen and Westbury, Andrew and Byrne, Eugene and Chavis, Zachary and Furnari, Antonino and Girdhar, Rohit and Hamburger, Jackson and Jiang, Hao and Liu, Miao and Liu, Xingyu and Martin, Miguel and Nagarajan, Tushar and Radosavovic, Ilija and Ramakrishnan, Santhosh Kumar and Ryan, Fiona and Sharma, Jayant and Wray, Michael and Xu, Mengmeng and Xu, Eric Zhongcong and Zhao, Chen and Bansal, Siddhant and Batra, Dhruv and Cartillier, Vincent and Crane, Sean and Do, Tien and Doulaty, Morrie and Erapalli, Akshay and Feichtenhofer, Christoph and Fragomeni, Adriano and Fu, Qichen and Gebreselasie, Abrham and Gonz\'alez, Cristina and Hillis, James and Huang, Xuhua and Huang, Yifei and Jia, Wenqi and Khoo, Weslie and Kol\'a\v{r}, J\'achym and Kottur, Satwik and Kumar, Anurag and Landini, Federico and Li, Chao and Li, Yanghao and Li, Zhenqiang and Mangalam, Karttikeya and Modhugu, Raghava and Munro, Jonathan and Murrell, Tullie and Nishiyasu, Takumi and Price, Will and Ruiz, Paola and Ramazanova, Merey and Sari, Leda and Somasundaram, Kiran and Southerland, Audrey and Sugano, Yusuke and Tao, Ruijie and Vo, Minh and Wang, Yuchen and Wu, Xindi and Yagi, Takuma and Zhao, Ziwei and Zhu, Yunyi and Arbel\'aez, Pablo and Crandall, David and Damen, Dima and Farinella, Giovanni Maria and Fuegen, Christian and Ghanem, Bernard and Ithapu, Vamsi Krishna and Jawahar, C. V. and Joo, Hanbyul and Kitani, Kris and Li, Haizhou and Newcombe, Richard and Oliva, Aude and Park, Hyun Soo and Rehg, James M. and Sato, Yoichi and Shi, Jianbo and Shou, Mike Zheng and Torralba, Antonio and Torresani, Lorenzo and Yan, Mingfei and Malik, Jitendra}, title = {Ego4D: Around the World in 3,000 Hours of Egocentric Video}, booktitle = {Proceedings of the IEEE/CVF Conference on Computer Vision and Pattern Recognition (CVPR)}, month = {June}, year = {2022}, pages = {18995-19012} }

@InProceedings{liu2025hoigen1mlargescaledatasethumanobject, author = {Liu, Kun and Liu, Qi and Liu, Xinchen and Li, Jie and Zhang, Yongdong and Luo, Jiebo and He, Xiaodong and Liu, Wu}, title = {HOIGen-1M: A Large-scale Dataset for Human-Object Interaction Video Generation}, booktitle = {Proceedings of the IEEE/CVF Conference on Computer Vision and Pattern Recognition (CVPR)}, month = {June}, year = {2025}, pages = {24001-24010} }

@misc{tan2024vidgen1mlargescaledatasettexttovideo,
      title={VidGen-1M: A Large-Scale Dataset for Text-to-video Generation},
      author={Zhiyu Tan and Xiaomeng Yang and Luozheng Qin and Hao Li},
      year={2024},
      eprint={2408.02629},
      archivePrefix={arXiv},
      primaryClass={cs.CV},
      url={https://arxiv.org/abs/2408.02629},
}

@inproceedings{
nan2025openvid1mlargescalehighqualitydataset,
title={OpenVid-1M: A Large-Scale High-Quality Dataset for Text-to-video Generation},
author={Kepan Nan and Rui Xie and Penghao Zhou and Tiehan Fan and Zhenheng Yang and Zhijie Chen and Xiang Li and Jian Yang and Ying Tai},
booktitle={The Thirteenth International Conference on Learning Representations},
year={2025},
url={https://openreview.net/forum?id=j7kdXSrISM}
}

@inproceedings{
mentzer2023finitescalarquantizationvqvae,
title={Finite Scalar Quantization: {VQ}-{VAE} Made Simple},
author={Fabian Mentzer and David Minnen and Eirikur Agustsson and Michael Tschannen},
booktitle={The Twelfth International Conference on Learning Representations},
year={2024},
url={https://openreview.net/forum?id=8ishA3LxN8}
}

@InProceedings{esser2024scalingrectifiedflowtransformers,
  title = 	 {Scaling Rectified Flow Transformers for High-Resolution Image Synthesis},
  author =       {Esser, Patrick and Kulal, Sumith and Blattmann, Andreas and Entezari, Rahim and M\"{u}ller, Jonas and Saini, Harry and Levi, Yam and Lorenz, Dominik and Sauer, Axel and Boesel, Frederic and Podell, Dustin and Dockhorn, Tim and English, Zion and Rombach, Robin},
  booktitle = 	 {Proceedings of the 41st International Conference on Machine Learning},
  pages = 	 {12606--12633},
  year = 	 {2024},
  editor = 	 {Salakhutdinov, Ruslan and Kolter, Zico and Heller, Katherine and Weller, Adrian and Oliver, Nuria and Scarlett, Jonathan and Berkenkamp, Felix},
  volume = 	 {235},
  series = 	 {Proceedings of Machine Learning Research},
  month = 	 {21--27 Jul},
  publisher =    {PMLR},
  url = 	 {https://proceedings.mlr.press/v235/esser24a.html}
}

@inproceedings{embodimentcollaboration2025openxembodimentroboticlearning, title={Open X-Embodiment: Robotic Learning Datasets and RT-X Models : Open X-Embodiment Collaboration}, url={http://dx.doi.org/10.1109/ICRA57147.2024.10611477}, DOI={10.1109/icra57147.2024.10611477}, booktitle={2024 IEEE International Conference on Robotics and Automation (ICRA)}, publisher={IEEE}, author={O’Neill, Abby and Rehman, Abdul and Maddukuri, Abhiram and Gupta, Abhishek and Padalkar, Abhishek and Lee, Abraham and Pooley, Acorn and Gupta, Agrim and Mandlekar, Ajay and Jain, Ajinkya and Tung, Albert and Bewley, Alex and Herzog, Alex and Irpan, Alex and Khazatsky, Alexander and Rai, Anant and Gupta, Anchit and Wang, Andrew and Singh, Anikait and Garg, Animesh and Kembhavi, Aniruddha and Xie, Annie and Brohan, Anthony and Raffin, Antonin and Sharma, Archit and Yavary, Arefeh and Jain, Arhan and Balakrishna, Ashwin and Wahid, Ayzaan and Burgess-Limerick, Ben and Kim, Beomjoon and Schölkopf, Bernhard and Wulfe, Blake and Ichter, Brian and Lu, Cewu and Xu, Charles and Le, Charlotte and Finn, Chelsea and Wang, Chen and Xu, Chenfeng and Chi, Cheng and Huang, Chenguang and Chan, Christine and Agia, Christopher and Pan, Chuer and Fu, Chuyuan and Devin, Coline and Xu, Danfei and Morton, Daniel and Driess, Danny and Chen, Daphne and Pathak, Deepak and Shah, Dhruv and Büchler, Dieter and Jayaraman, Dinesh and Kalashnikov, Dmitry and Sadigh, Dorsa and Johns, Edward and Foster, Ethan and Liu, Fangchen and Ceola, Federico and Xia, Fei and Zhao, Feiyu and Stulp, Freek and Zhou, Gaoyue and Sukhatme, Gaurav S. and Salhotra, Gautam and Yan, Ge and Feng, Gilbert and Schiavi, Giulio and Berseth, Glen and Kahn, Gregory and Wang, Guanzhi and Su, Hao and Fang, Hao-Shu and Shi, Haochen and Bao, Henghui and Ben Amor, Heni and Christensen, Henrik I and Furuta, Hiroki and Walke, Homer and Fang, Hongjie and Ha, Huy and Mordatch, Igor and Radosavovic, Ilija and Leal, Isabel and Liang, Jacky and Abou-Chakra, Jad and Kim, Jaehyung and Drake, Jaimyn and Peters, Jan and Schneider, Jan and Hsu, Jasmine and Bohg, Jeannette and Bingham, Jeffrey and Wu, Jeffrey and Gao, Jensen and Hu, Jiaheng and Wu, Jiajun and Wu, Jialin and Sun, Jiankai and Luo, Jianlan and Gu, Jiayuan and Tan, Jie and Oh, Jihoon and Wu, Jimmy and Lu, Jingpei and Yang, Jingyun and Malik, Jitendra and Silvério, João and Hejna, Joey and Booher, Jonathan and Tompson, Jonathan and Yang, Jonathan and Salvador, Jordi and Lim, Joseph J. and Han, Junhyek and Wang, Kaiyuan and Rao, Kanishka and Pertsch, Karl and Hausman, Karol and Go, Keegan and Gopalakrishnan, Keerthana and Goldberg, Ken and Byrne, Kendra and Oslund, Kenneth and Kawaharazuka, Kento and Black, Kevin and Lin, Kevin and Zhang, Kevin and Ehsani, Kiana and Lekkala, Kiran and Ellis, Kirsty and Rana, Krishan and Srinivasan, Krishnan and Fang, Kuan and Singh, Kunal Pratap and Zeng, Kuo-Hao and Hatch, Kyle and Hsu, Kyle and Itti, Laurent and Chen, Lawrence Yunliang and Pinto, Lerrel and Fei-Fei, Li and Tan, Liam and Fan, Linxi Jim and Ott, Lionel and Lee, Lisa and Weihs, Luca and Chen, Magnum and Lepert, Marion and Memmel, Marius and Tomizuka, Masayoshi and Itkina, Masha and Castro, Mateo Guaman and Spero, Max and Du, Maximilian and Ahn, Michael and Yip, Michael C. and Zhang, Mingtong and Ding, Mingyu and Heo, Minho and Srirama, Mohan Kumar and Sharma, Mohit and Kim, Moo Jin and Kanazawa, Naoaki and Hansen, Nicklas and Heess, Nicolas and Joshi, Nikhil J and Suenderhauf, Niko and Liu, Ning and Di Palo, Norman and Shafiullah, Nur Muhammad Mahi and Mees, Oier and Kroemer, Oliver and Bastani, Osbert and Sanketi, Pannag R and Miller, Patrick Tree and Yin, Patrick and Wohlhart, Paul and Xu, Peng and Fagan, Peter David and Mitrano, Peter and Sermanet, Pierre and Abbeel, Pieter and Sundaresan, Priya and Chen, Qiuyu and Vuong, Quan and Rafailov, Rafael and Tian, Ran and Doshi, Ria and Martín-Martín, Roberto and Baijal, Rohan and Scalise, Rosario and Hendrix, Rose and Lin, Roy and Qian, Runjia and Zhang, Ruohan and Mendonca, Russell and Shah, Rutav and Hoque, Ryan and Julian, Ryan and Bustamante, Samuel and Kirmani, Sean and Levine, Sergey and Lin, Shan and Moore, Sherry and Bahl, Shikhar and Dass, Shivin and Sonawani, Shubham and Song, Shuran and Xu, Sichun and Haldar, Siddhant and Karamcheti, Siddharth and Adebola, Simeon and Guist, Simon and Nasiriany, Soroush and Schaal, Stefan and Welker, Stefan and Tian, Stephen and Ramamoorthy, Subramanian and Dasari, Sudeep and Belkhale, Suneel and Park, Sungjae and Nair, Suraj and Mirchandani, Suvir and Osa, Takayuki and Gupta, Tanmay and Harada, Tatsuya and Matsushima, Tatsuya and Xiao, Ted and Kollar, Thomas and Yu, Tianhe and Ding, Tianli and Davchev, Todor and Zhao, Tony Z. and Armstrong, Travis and Darrell, Trevor and Chung, Trinity and Jain, Vidhi and Vanhoucke, Vincent and Zhan, Wei and Zhou, Wenxuan and Burgard, Wolfram and Chen, Xi and Wang, Xiaolong and Zhu, Xinghao and Geng, Xinyang and Liu, Xiyuan and Liangwei, Xu and Li, Xuanlin and Lu, Yao and Ma, Yecheng Jason and Kim, Yejin and Chebotar, Yevgen and Zhou, Yifan and Zhu, Yifeng and Wu, Yilin and Xu, Ying and Wang, Yixuan and Bisk, Yonatan and Cho, Yoonyoung and Lee, Youngwoon and Cui, Yuchen and Cao, Yue and Wu, Yueh-Hua and Tang, Yujin and Zhu, Yuke and Zhang, Yunchu and Jiang, Yunfan and Li, Yunshuang and Li, Yunzhu and Iwasawa, Yusuke and Matsuo, Yutaka and Ma, Zehan and Xu, Zhuo and Cui, Zichen Jeff and Zhang, Zichen and Lin, Zipeng}, year={2024}, month=may, pages={6892–6903} }

@INPROCEEDINGS{khazatsky2025droidlargescaleinthewildrobot,
    AUTHOR    = {Alexander Khazatsky AND Karl Pertsch AND Suraj Nair AND Ashwin Balakrishna AND Sudeep Dasari AND Siddharth Karamcheti AND Soroush Nasiriany AND Mohan Kumar Srirama AND Lawrence Yunliang Chen AND Kirsty Ellis AND Peter David Fagan AND Joey Hejna AND Masha Itkina AND Marion Lepert AND Yecheng Jason Ma AND Patrick Tree Miller AND Jimmy Wu AND Suneel Belkhale AND Shivin Dass AND Huy Ha AND Arhan Jain AND Abraham Lee AND Youngwoon Lee AND Marius Memmel AND Sungjae Park AND Ilija Radosavovic AND Kaiyuan Wang AND Albert Zhan AND Kevin Black AND Cheng Chi AND Kyle Beltran Hatch AND Shan Lin AND Jingpei Lu AND Jean Mercat AND Abdul Rehman AND Pannag R Sanketi AND Archit Sharma AND Cody Simpson AND Quan Vuong AND Homer Rich Walke AND Blake Wulfe AND Ted Xiao AND Jonathan Heewon Yang AND Arefeh Yavary AND Tony Z. Zhao AND Christopher Agia AND Rohan Baijal AND Mateo Guaman Castro AND Daphne Chen AND Qiuyu Chen AND Trinity Chung AND Jaimyn Drake AND Ethan Paul Foster AND Jensen Gao AND David Antonio Herrera AND Minho Heo AND Kyle Hsu AND Jiaheng Hu AND Donovon Jackson AND Charlotte Le AND Yunshuang Li AND Roy Lin AND Zehan Ma AND Abhiram Maddukuri AND Suvir Mirchandani AND Daniel Morton AND Tony Nguyen AND Abigail O'Neill AND Rosario Scalise AND Derick Seale AND Victor Son AND Stephen Tian AND Emi Tran AND Andrew E. Wang AND Yilin Wu AND Annie Xie AND Jingyun Yang AND Patrick Yin AND Yunchu Zhang AND Osbert Bastani AND Glen Berseth AND Jeannette Bohg AND Ken Goldberg AND Abhinav Gupta AND Abhishek Gupta AND Dinesh Jayaraman AND Joseph J Lim AND Jitendra Malik AND Roberto Martín-Martín AND Subramanian Ramamoorthy AND Dorsa Sadigh AND Shuran Song AND Jiajun Wu AND Michael C. Yip AND Yuke Zhu AND Thomas Kollar AND Sergey Levine AND Chelsea Finn},
    TITLE     = {{DROID: A Large-Scale In-The-Wild Robot Manipulation Dataset}},
    BOOKTITLE = {Proceedings of Robotics: Science and Systems},
    YEAR      = {2024},
    ADDRESS   = {Delft, Netherlands},
    MONTH     = {July},
    DOI       = {10.15607/RSS.2024.XX.120}
}

@INPROCEEDINGS{black2026pi0visionlanguageactionflowmodel,
    AUTHOR    = {Kevin Black AND Noah Brown AND Danny Driess AND Adnan Esmail AND Michael Robert Equi AND Chelsea Finn AND Niccolo Fusai AND Lachy Groom AND Karol Hausman AND Brian Ichter AND Szymon Jakubczak AND Tim Jones AND Liyiming Ke AND Sergey Levine AND Adrian Li-Bell AND Mohith Mothukuri AND Suraj Nair AND Karl Pertsch AND Lucy Xiaoyang Shi AND Laura Smith AND James Tanner AND Quan Vuong AND Anna Walling AND Haohuan Wang AND Ury Zhilinsky},
    TITLE     = {{{$\pi_0$}: A Vision-Language-Action Flow Model for General Robot Control}},
    BOOKTITLE = {Proceedings of Robotics: Science and Systems},
    YEAR      = {2025},
    ADDRESS   = {LosAngeles, CA, USA},
    MONTH     = {June},
    DOI       = {10.15607/RSS.2025.XXI.010}
}

@InProceedings{wang2023videomaev2scalingvideo, author = {Wang, Limin and Huang, Bingkun and Zhao, Zhiyu and Tong, Zhan and He, Yinan and Wang, Yi and Wang, Yali and Qiao, Yu}, title = {VideoMAE V2: Scaling Video Masked Autoencoders With Dual Masking}, booktitle = {Proceedings of the IEEE/CVF Conference on Computer Vision and Pattern Recognition (CVPR)}, month = {June}, year = {2023}, pages = {14549-14560} }

@misc{agibotworldcontributors2025agibotworldcolosseolargescale,
      title={AgiBot World Colosseo: A Large-scale Manipulation Platform for Scalable and Intelligent Embodied Systems},
      author={AgiBot-World-Contributors and Qingwen Bu and Jisong Cai and Li Chen and Xiuqi Cui and Yan Ding and Siyuan Feng and Shenyuan Gao and Xindong He and Xuan Hu and Xu Huang and Shu Jiang and Yuxin Jiang and Cheng Jing and Hongyang Li and Jialu Li and Chiming Liu and Yi Liu and Yuxiang Lu and Jianlan Luo and Ping Luo and Yao Mu and Yuehan Niu and Yixuan Pan and Jiangmiao Pang and Yu Qiao and Guanghui Ren and Cheng Ruan and Jiaqi Shan and Yongjian Shen and Chengshi Shi and Mingkang Shi and Modi Shi and Chonghao Sima and Jianheng Song and Huijie Wang and Wenhao Wang and Dafeng Wei and Chengen Xie and Guo Xu and Junchi Yan and Cunbiao Yang and Lei Yang and Shukai Yang and Maoqing Yao and Jia Zeng and Chi Zhang and Qinglin Zhang and Bin Zhao and Chengyue Zhao and Jiaqi Zhao and Jianchao Zhu},
      year={2025},
      eprint={2503.06669},
      archivePrefix={arXiv},
      primaryClass={cs.RO},
      url={https://arxiv.org/abs/2503.06669},
}

@misc{beattie2016deepmindlab,
      title={DeepMind Lab},
      author={Charles Beattie and Joel Z. Leibo and Denis Teplyashin and Tom Ward and Marcus Wainwright and Heinrich Küttler and Andrew Lefrancq and Simon Green and Víctor Valdés and Amir Sadik and Julian Schrittwieser and Keith Anderson and Sarah York and Max Cant and Adam Cain and Adrian Bolton and Stephen Gaffney and Helen King and Demis Hassabis and Shane Legg and Stig Petersen},
      year={2016},
      eprint={1612.03801},
      archivePrefix={arXiv},
      primaryClass={cs.AI},
      url={https://arxiv.org/abs/1612.03801},
}

@inproceedings{
liu2022flowstraightfastlearning,
title={Flow Straight and Fast: Learning to Generate and Transfer Data with Rectified Flow},
author={Xingchao Liu and Chengyue Gong and qiang liu},
booktitle={The Eleventh International Conference on Learning Representations },
year={2023},
url={https://openreview.net/forum?id=XVjTT1nw5z}
}

@misc{liu2026oatorderedactiontokenization,
      title={OAT: Ordered Action Tokenization},
      author={Chaoqi Liu and Xiaoshen Han and Jiawei Gao and Yue Zhao and Haonan Chen and Yilun Du},
      year={2026},
      eprint={2602.04215},
      archivePrefix={arXiv},
      primaryClass={cs.RO},
      url={https://arxiv.org/abs/2602.04215},
}

@inproceedings{
shen2025catcontentadaptiveimagetokenization,
title={{CAT}: Content-Adaptive Image Tokenization},
author={Junhong Shen and Kushal Tirumala and Michihiro Yasunaga and Ishan Misra and Luke Zettlemoyer and LILI YU and Chunting Zhou},
booktitle={The Thirty-ninth Annual Conference on Neural Information Processing Systems},
year={2026},
url={https://openreview.net/forum?id=cot6mZPkWo}
}

@inproceedings{
lipman2023flowmatchinggenerativemodeling,
title={Flow Matching for Generative Modeling},
author={Yaron Lipman and Ricky T. Q. Chen and Heli Ben-Hamu and Maximilian Nickel and Matthew Le},
booktitle={The Eleventh International Conference on Learning Representations },
year={2023},
url={https://openreview.net/forum?id=PqvMRDCJT9t}
}

@InProceedings{zhang2018unreasonableeffectivenessdeepfeatures,
author = {Zhang, Richard and Isola, Phillip and Efros, Alexei A. and Shechtman, Eli and Wang, Oliver},
title = {The Unreasonable Effectiveness of Deep Features as a Perceptual Metric},
booktitle = {Proceedings of the IEEE Conference on Computer Vision and Pattern Recognition (CVPR)},
month = {June},
year = {2018}
}
}

%%%%%%%%%%%%%%%%%%%%%%%%%%%%%%%%%%%%%%%%%%%%%%%%%%%%%%%%%%%%%%%%%%%%%%%%%%%%%%%
%%%%%%%%%%%%%%%%%%%%%%%%%%%%%%%%%%%%%%%%%%%%%%%%%%%%%%%%%%%%%%%%%%%%%%%%%%%%%%%
% APPENDIX
%%%%%%%%%%%%%%%%%%%%%%%%%%%%%%%%%%%%%%%%%%%%%%%%%%%%%%%%%%%%%%%%%%%%%%%%%%%%%%%
%%%%%%%%%%%%%%%%%%%%%%%%%%%%%%%%%%%%%%%%%%%%%%%%%%%%%%%%%%%%%%%%%%%%%%%%%%%%%%%
\newpage
\appendix

\section{DMLab Experimental Details}
\label{app:dmlab_details}

This appendix provides the experimental details needed to reproduce the DMLab
experiments. We describe the environments, expert-video collection, task filtering,
bottleneck configurations, architectures, decoder objective, training stages, and
prefix-length sampling. The goal is to make clear that the main DMLab comparisons
isolate the bottleneck design because the surrounding LAM, translator, and
latent-token sequence model are kept fixed across methods whenever possible.

\subsection{Environments and Expert Video Dataset}
\label{app:env-expert-video}

We evaluate downstream task performance in DeepMind Lab (DMLab)
~\citep{beattie2016deepmindlab}. DMLab provides egocentric partially observed
environments with viewpoint changes, occlusions, distractors, and task-dependent
visual structure. These properties make it a useful testbed for studying latent-action
representations because the observation transition $(o_t,o_{t+1})$ contains both
potentially action-relevant changes and nuisance variation.

Expert trajectories are collected by rolling out agents trained with DreamerV3
~\citep{hafner2024masteringdiversedomainsworld}. Observations are RGB images of size
$64\times64$. We use the resulting trajectories in two forms. The action-free portion
provides observation transitions $(o_t,o_{t+1})$ for LAM pretraining and latent-token
sequence-model training. A much smaller action-labeled subset provides
$(o_t,o_{t+1},a_t)$ tuples for translator training and evaluation.

For each environment in Table~\ref{tab:dmlab_envs}, the dataset contains 9,000,000
recorded training steps and 1,000,000 recorded test steps. The DreamerV3 returns in
the table are used to normalize downstream task performance in the main text and in
Appendix~\ref{app:dmlab_results}.

\begin{table}[ht]
\centering
\caption{\textbf{DreamerV3 expert returns.}
Returns for the DMLab environments used to normalize downstream task performance.}
\label{tab:dmlab_envs}
\begin{tabular}{lr}
\toprule
Environment & Return \\
\midrule
explore\_goal\_locations\_large          & 158.82 \\
explore\_goal\_locations\_small          & 368.09 \\
explore\_object\_locations\_large        &  56.12 \\
explore\_object\_locations\_small        &  92.00 \\
explore\_object\_rewards\_few            &  40.83 \\
explore\_object\_rewards\_many           &  53.45 \\
explore\_obstructed\_goals\_large        &  60.02 \\
explore\_obstructed\_goals\_small        & 269.45 \\
language\_execute\_random\_task          & -10.14 \\
lasertag\_one\_opponent\_large           &  -0.03 \\
lasertag\_one\_opponent\_small           &  -0.06 \\
lasertag\_three\_opponent\_large         &   7.40 \\
natlab\_varying\_map\_regrowth           &   9.49 \\
psychlab\_visual\_search                 &  39.85 \\
rooms\_exploit\_deferred\_effects\_train &  40.16 \\
rooms\_watermaze                         &  28.32 \\
\bottomrule
\end{tabular}
\end{table}

\subsection{Task Filtering and Normalization}
\label{app:task_filtering}

Expert trajectories are collected for all 16 DMLab tasks. For normalized-return
evaluation, we exclude five tasks with extremely low or unstable expert returns:
Language Execute Random Task, Lasertag One Opponent Large, Lasertag One Opponent
Small, Lasertag Three Opponent Large, and Natlab Varying Map Regrowth. The
remaining 11 tasks are used for reported normalized returns.

For each task, normalized return is computed as a percentage of the DreamerV3 expert
return. This makes scores comparable across tasks with different reward scales. The
resulting averages support controlled comparisons among bottleneck designs under
scarce or narrowly distributed action-alignment labels.

The single-task alignment source, Lasertag One Opponent Large, is included in the
available trajectory collection but excluded from the normalized evaluation suite.

\subsection{Bottleneck Settings for FlexLAM and Fixed-K Baselines}
\label{app:bottleneck_settings}

The main DMLab comparison uses controlled Fixed-K baselines rather than only the
previous tight/loose endpoint comparison. Fixed-K$k$ models are trained separately
with a fixed $k$-token bottleneck. FlexLAM uses the same maximum $K=64$ code space
and is evaluated as FlexLAM@$k$ by retaining the first $k$ tokens. The latent
sequence is quantized with FSQ~\citep{mentzer2023finitescalarquantizationvqvae}.

\begin{table}[ht]
\centering
\small
\setlength{\tabcolsep}{4pt}
\caption{\textbf{Bottleneck settings for DMLab.}
Fixed-K$k$ models are trained separately at each token budget. FlexLAM is trained
once with retained-prefix sampling and evaluated at multiple prefix lengths.}
\label{tab:bottleneck_settings}
\begin{tabularx}{\linewidth}{@{}lcccX@{}}
\toprule
Method & FSQ levels & Train $K$ & Retained-prefix & Notes \\
\midrule
Fixed-K2$^\dagger$  & [5] & 2 & $\times$ & fixed-capacity baseline trained separately \\
Fixed-K4  & [8, 5, 5, 5] & 4  & $\times$ & fixed-capacity baseline trained separately \\
Fixed-K16 & [8, 5, 5, 5] & 16 & $\times$ & fixed-capacity baseline trained separately \\
Fixed-K64 & [8, 5, 5, 5] & 64 & $\times$ & loose fixed-capacity endpoint \\
FlexLAM   & [8, 5, 5, 5] & 64 & \checkmark & trained once; evaluated as FlexLAM@$k$ \\
\bottomrule
\end{tabularx}
\end{table}

\begin{table}[t]
\centering
\caption{\textbf{Nominal discrete bottleneck capacity in DMLab.}
Fixed-K$k$ and FlexLAM@$k$ use the same FSQ vocabulary at the same evaluated token
budget. Capacity is computed as $k\log_2|\mathcal{C}|$, where
$|\mathcal{C}|=\prod_i L_i$ for FSQ levels $\mathcal{L}$.}
\label{tab:dmlab_nominal_capacity}
\begin{tabular}{lccc}
\toprule
Operating point & Quantizer & Tokens & Nominal capacity \\
\midrule
Fixed-K4 / FlexLAM@4   & FSQ [8,5,5,5] & 4  & $4\log_2 1000 \approx 39.9$ bits \\
Fixed-K16 / FlexLAM@16 & FSQ [8,5,5,5] & 16 & $16\log_2 1000 \approx 159.5$ bits \\
Fixed-K64 / FlexLAM@64 & FSQ [8,5,5,5] & 64 & $64\log_2 1000 \approx 637.8$ bits \\
\bottomrule
\end{tabular}
\end{table}

\subsection{LAM and Translator Architectures}
\label{app:dmlab_architectures}

We keep the encoder, decoder, and translator architectures fixed across methods
whenever possible. Only the bottleneck configuration differs. The encoder processes
two consecutive frames and outputs a token sequence. The decoder is conditioned on
$o_t$ and the null-filled retained-prefix representation $\tilde{z}_t^{(k)}$ to decode
the transition target under the decoder objective. The decoder follows a DiT-style
architecture~\citep{peebles2023scalablediffusionmodelstransformers}. DMLab LAM hyperparameters are listed in
Table~\ref{tab:dmlab_lam_hparams}.

\begin{table}[ht]
\centering
\caption{\textbf{DMLab LAM hyperparameters.}
Key architectural settings for the encoder and decoder.}
\label{tab:dmlab_lam_hparams}
\begin{tabular}{lll}
\toprule
Component & Parameter & Value \\
\midrule
Encoder & depth & 8 \\
& embed dim & 192 \\
& mlp ratio & 4 \\
& tubelet size & 2 \\
& patch size & $4 \times 4$ \\
\midrule
Decoder (DiT) & num layers & 12 \\
& num heads & 3 \\
& head dim & 64 \\
& ffn dim & 776 \\
& patch size & $4 \times 4$ \\
\bottomrule
\end{tabular}
\end{table}

The translator maps latent transition tokens to executable actions. We use an
Hourglass MLP~\citep{chen2025rethinkingshapeconventionmlp}. Discrete action dimensions are trained
with cross-entropy; continuous dimensions, when present, are trained with MSE. The
translator conditions on the previous action $a_{t-1}$ in all compared methods. This
conditioning choice is shared by FlexLAM and the fixed-capacity baselines.

\begin{table}[ht]
\centering
\caption{\textbf{Translator hyperparameters.}}
\label{tab:translator_hparams}
\begin{tabular}{lr}
\toprule
Parameter & Value \\
\midrule
depth & 5 \\
hidden dim & 96 \\
wide dim & 512 \\
\bottomrule
\end{tabular}
\end{table}

\subsection{Decoder Objective Details}
\label{app:dmlab_decoder_objective}

Both DMLab and real-world decoders are trained with a rectified-flow objective
~\citep{lipman2023flowmatchinggenerativemodeling,liu2022flowstraightfastlearning,esser2024scalingrectifiedflowtransformers}. In both settings,
the decoder is conditioned on the current observation $o_t$ and the null-filled
retained-prefix representation $\tilde{z}_t^{(k)}$. The retained-prefix conditioning
principle is therefore shared across simulated and real-world video experiments, even
though the two settings differ in architecture, resolution, initialization, and data
mixture.

For DMLab, the encoder and decoder are trained from scratch on $64\times64$ RGB
observation transitions. For real-world video, the encoder and decoder are initialized
from pretrained video and image-generation models, as described in
Appendix~\ref{app:real_world_details}. We use $\mathcal{L}_{\mathrm{dec}}$ in the
main text to denote this transition-decoding objective abstractly, since the same
retained-prefix bottleneck is used across settings while the decoder family and data
regime differ.

\subsection{Training Stages and Checkpoint Selection}
\label{app:dmlab_training}

DMLab training uses three main stages. An optional fourth stage is used for the
joint-alignment variant in Section~\ref{subsec:joint_training}. Each bottleneck design
uses a single NVIDIA H100 GPU; LAM pretraining takes approximately 24 hours per
method, and latent-token sequence-model training takes approximately 46 hours per
method.

\paragraph{Stage 1. LAM pretraining.}
We train the encoder and decoder on action-free transitions
$\mathcal{D}_u=\{(o_t,o_{t+1})\}$. FlexLAM samples a retained prefix length and
conditions the decoder on the null-filled retained-prefix representation
$\tilde{z}_t^{(k)}$. Fixed-capacity baselines expose their fixed bottleneck to the
decoder.

\paragraph{Stage 2. latent-token sequence-model training.}
We train the latent-token sequence model on trajectories represented as interleaved
continuous observation patch embeddings and latent-action code blocks.
This stage uses only action-free data. The sequence-model architecture is fixed
across methods, but the latent codes come from each method's LAM.

\paragraph{Stage 3. translator training.}
We train the translator on the small labeled subset
$\mathcal{D}_e=\{(o_t,o_{t+1},a_t)\}$. Unless otherwise stated, the LAM and
latent-token sequence model are frozen. We select checkpoints using translator
validation loss, which correlates with downstream task performance in our DMLab
experiments.

\paragraph{Optional stage. joint alignment.}
Section~\ref{subsec:joint_training} evaluates a performance-oriented joint-alignment
variant where the action loss is allowed to update the LAM encoder and bottleneck
parameters together with the translator.

\subsection{Prefix-Length Sampling Distribution}
\label{app:prefix_length_distribution}

In all stages that use prefix sampling, we sample
$k\sim\mathrm{Unif}\{0,\ldots,K\}$ independently for each example. Suffix slots
$j>k$ are replaced by the shared learnable null latent. The $k=0$ case is used only
during training and corresponds to an all-null latent input; all reported operating
points use $k>0$. Biasing $p(k)$ toward shorter prefixes may encourage more
aggressive compression, while biasing it toward longer prefixes may improve
high-capacity decoding.

%%%%%%%%%%%%%%%%%%%%%%%%%%%%%%%%%%%%%%%%%%%%%%%%%%%%%%%%%%%%%%%%%%%%%%%%%%%%%%%

\section{Latent-Token Sequence Model and Inference Details}
\label{app:sequence_model}

The main text treats the latent-token sequence model as a downstream evaluator. We
include the details here to make the DMLab rollout procedure reproducible. This
component determines how latent actions are generated during DMLab evaluation,
whereas FlexLAM changes how the latent-action bottleneck is trained. This separation
keeps the experiments focused on the bottleneck comparison with a shared sequence
model.

\subsection{Sequence Construction}
\label{app:sequence_construction}

For each prediction window, frames are maintained in a 34-frame context and
subsampled at stride $s=8$, while the latest frame is always included. A typical
context for predicting the next latent-action block has the form
\[
[\ldots, x_{t-16},
c_{t-16,1:K},\ldots,c_{t-9,1:K},
x_{t-8},
c_{t-8,1:K},\ldots,c_{t-1,1:K},
x_t].
\]
Here, $x_t$ denotes continuous observation patch embeddings. Each
$c_{u,1:K}$ is a latent-action code block encoded from the observed
transition $(o_u,o_{u+1})$. The same 34-frame context length and latest-frame
refresh pattern are used during inference.

\subsection{Sequence-Model Training Targets}
\label{app:sequence_model_targets}

The latent-token sequence model is trained on latent-action code blocks
encoded from observed transitions. Prefix truncation is used for LAM/translator
training and for current-step action selection at inference; sequence-model training
targets remain complete code blocks. Observation patch embeddings serve as
conditioning inputs.

The sequence model minimizes next-token prediction loss over latent-action code
positions. Let $\mathcal{I}_{\mathrm{LA}}$ denote positions corresponding to
latent-action codes. The objective is
\[
\min_{\omega}
\mathbb{E}_{\mathbf{y}\sim\mathcal{D}_u}
\left[
\sum_{i\in\mathcal{I}_{\mathrm{LA}}}
-\log p_\omega(c_i\mid \mathbf{y}_{<i})
\right].
\]

\subsection{Translator Training Token Source}
\label{app:translator_token_source}

During translator training, latent codes are obtained from the LAM encoder applied
to observed transitions. During rollout, the translator
receives sequence-model-generated codes for the current transition. This follows the
standard LAM pipeline in which the translator learns the latent-to-action map from
encoded transitions, while the sequence model supplies predicted latent actions at
decision time.

\subsection{Sequence Model Architecture and Hyperparameters}
\label{app:sequence_model_hparams}

The evaluator is a decoder-only causal transformer with RoPE and a Qwen-style
configuration. Table~\ref{tab:sequence_model_hparams} lists the architecture,
training curriculum, context length, and inference-cache settings. These settings are
shared across bottleneck designs so that downstream differences reflect the
latent-action codes under a common sequence-model architecture.

\begin{table}[ht]
\centering
\caption{\textbf{Latent-token sequence model hyperparameters.}
These settings correspond to the downstream sequence model used in DMLab.}
\label{tab:sequence_model_hparams}
\begin{tabular}{lr}
\toprule
Parameter & Value \\
\midrule
architecture & Qwen3-style decoder-only transformer \\
positional encoding & RoPE \\
hidden size & 256 \\
intermediate size & 1024 \\
num attention heads & 4 \\
num key-value heads & 4 \\
image patch size & 8 \\
image tokens per frame & 64 \\
input frame stride & 8 \\
total training steps & 300k \\
LR schedule & cosine decay \\
curriculum stage1 num\_frames & 10 \\
curriculum stage1 batch size & 256 \\
curriculum stage1 steps & $\sim$250k \\
curriculum stage2 num\_frames & 34 \\
curriculum stage2 batch size & 128 \\
curriculum stage2 steps & 250k--300k \\
context frames at inference & 34 \\
latency benchmark context frames & 34 \\
KV cache at inference & enabled \\
\bottomrule
\end{tabular}
\end{table}

\subsection{Inference Procedure}
\label{app:inference}

Algorithm~\ref{alg:flexlam_inference} gives one decision step of the DMLab rollout
procedure. The sequence model generates only the current
latent-action prefix; after the next observation arrives, the realized transition is
encoded and appended to the history. Reducing $k$
shortens current-step generation while the stored history remains unchanged. For
fixed-capacity baselines, $k$ is fixed to the bottleneck size.

\begin{algorithm}[ht]
\caption{One FlexLAM decision step with sparse observation context}
\label{alg:flexlam_inference}
\begin{algorithmic}[1]
\REQUIRE observations $o_{t-1}, o_t$, previous action $a_{t-1}$
\REQUIRE frame buffer $\mathcal{Q}$, latent-code buffer $\mathcal{B}$
\REQUIRE code encoder $F_\theta$, sequence model $p_\omega$, translator $g_\psi$
\REQUIRE observation stride $s$, retained prefix length $k$
\ENSURE predicted action $\hat{a}_t$
\IF{$t>0$}
    \STATE $c_{t-1,1:K} \leftarrow F_\theta(o_{t-1}, o_t)$
    \STATE $\mathcal{B} \leftarrow \mathrm{Append}(\mathcal{B}, c_{t-1,1:K})$
\ENDIF
\STATE $\mathcal{Q} \leftarrow \mathrm{Append}(\mathcal{Q}, o_t)$
\STATE $\mathcal{H}_t \leftarrow \mathrm{BuildContext}(\mathcal{Q}, \mathcal{B}; s)$
\STATE $\hat{c}_{t,1:k} \leftarrow \mathrm{DecodePrefix}(p_\omega,\mathcal{H}_t,k)$
\STATE $\tilde{z}_{t}^{(k)} \leftarrow \mathrm{NullFill}(\hat{c}_{t,1:k})$
\STATE $\hat{a}_t \leftarrow g_\psi(\tilde{z}_{t}^{(k)}, a_{t-1})$
\STATE \textbf{return} $\hat{a}_t$
\end{algorithmic}
\end{algorithm}

Here, $F_\theta$ denotes the LAM encoder and quantizer composed as a code encoder,
so $F_\theta(o_{t-1},o_t)$ returns discrete latent-action codes
$c_{t-1,1:K}$. $\mathrm{BuildContext}$ constructs sparse observation embeddings
from $\mathcal{Q}$ using stride $s$ and interleaves them with cached latent-action
codes from $\mathcal{B}$. $\mathrm{DecodePrefix}$ applies the latent-token sequence
model autoregressively for $k$ code positions. $\mathrm{NullFill}$ embeds predicted
codes into latent vectors and fills suffix slots with the shared null latent.

\subsection{Latency Measurement Protocol}
\label{app:latency_measurement}

Latency in Table~\ref{tab:token_length_analysis} is measured as wall-clock time per
decision step using the same 34-frame context length used at inference and in the
second stage of sequence-model training. Measurements use a single NVIDIA RTX 6000
Ada GPU, batch size 1, bf16 precision, and KV cache enabled. Each decision step
includes image preprocessing, encoding the newest observed transition
$(o_{t-1},o_t)$ into latent-action codes, sequence-model context
construction, autoregressive generation of $k$ current latent-action codes, null
filling, and action decoding. Previously encoded history tokens are cached and are
not recomputed. Reported values are means over 100 steady-state decision steps after
20 warmup steps.

\subsection{DMLab Latent-Token Prediction Visualization}
\label{app:latent_token_prediction}

This visualization documents the behavior of the downstream evaluation pipeline.

\begin{figure}[ht]
  \centering
  \includegraphics[width=\linewidth]{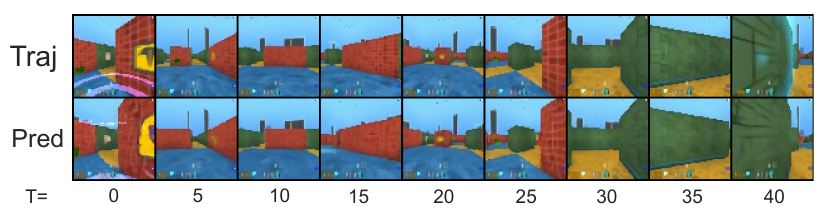}
  \caption{\textbf{DMLab latent-token prediction visualization.}
  We decode latent tokens generated by the downstream sequence model to visualize
  predicted one-step transitions and illustrate the behavior of the evaluation pipeline.}
  \label{fig:latent_token_prediction_app}
\end{figure}

%%%%%%%%%%%%%%%%%%%%%%%%%%%%%%%%%%%%%%%%%%%%%%%%%%%%%%%%%%%%%%%%%%%%%%%%%%%%%%%

\section{Full DMLab Results}
\label{app:dmlab_results}

\subsection{Per-Task Normalized Returns}
\label{app:dmlab_normalized_returns}

Table~\ref{tab:dmlab_normalized_scores} reports per-task normalized returns under
both standard scarce-label supervision and action alignment from a narrow single-task
source. Standard supervision uses 0.025\% labels sampled uniformly across tasks. The
biased columns use labels from Lasertag One Opponent Large.

We report mean $\pm$ standard error over 50 evaluation episodes per task. The table includes the tight
Fixed-K endpoint, the full-capacity Fixed-K endpoint, and the corresponding FlexLAM
operating points used in the main analysis.

\begin{table}[ht]
\centering
\small
\setlength{\tabcolsep}{4pt}
\renewcommand{\arraystretch}{1.05}
\caption{\textbf{Per-task normalized returns.}
Downstream task return normalized by DreamerV3 expert performance under standard
scarce-label supervision (0.025\% labels) and action alignment from a narrow
single-task source. Columns marked biased use labels from Lasertag One Opponent
Large; the source task is excluded from the normalized evaluation suite. We report
mean $\pm$ standard error over 50 evaluation episodes per task. $^\dagger$ denotes
the two-token tight fixed-capacity endpoint.}
\label{tab:dmlab_normalized_scores}
\resizebox{\textwidth}{!}{%
\begin{tabular}{lccccccc}
\toprule
Task & Random & Fixed-K2$^\dagger$ & FlexLAM@1 & Fixed-K64 & FlexLAM@64
& \shortstack{Fixed-K64\\(biased)} & \shortstack{FlexLAM@64\\(biased)} \\
\midrule
explore\_goal\_locations\_large
& 1.95 & 5.20\Std{1.27} & 13.00\Std{1.98}
& 14.60\Std{2.14} & 15.40\Std{2.53}
& 6.60\Std{1.46} & 9.80\Std{1.48} \\
explore\_goal\_locations\_small
& 2.09 & 15.40\Std{2.39} & 27.60\Std{3.27}
& 30.20\Std{2.99} & 38.40\Std{4.70}
& 24.60\Std{3.02} & 19.20\Std{3.16} \\
explore\_object\_locations\_large
& 8.37 & 7.56\Std{0.44} & 12.00\Std{0.46}
& 11.98\Std{0.48} & 14.10\Std{0.62}
& 8.72\Std{0.41} & 10.58\Std{0.43} \\
explore\_object\_locations\_small
& 3.91 & 5.46\Std{0.24} & 11.04\Std{0.52}
& 9.76\Std{0.65} & 12.42\Std{0.89}
& 7.06\Std{0.47} & 8.54\Std{0.71} \\
explore\_object\_rewards\_few
& 5.14 & 1.66\Std{0.32} & 6.02\Std{0.73}
& 6.86\Std{0.73} & 10.48\Std{0.96}
& 5.44\Std{0.75} & 7.94\Std{1.21} \\
explore\_object\_rewards\_many
& 4.49 & 4.14\Std{0.56} & 10.20\Std{1.09}
& 10.74\Std{0.85} & 16.08\Std{1.45}
& 8.00\Std{0.80} & 10.22\Std{1.08} \\
explore\_obstructed\_goals\_large
& 4.33 & 5.80\Std{1.33} & 10.60\Std{1.80}
& 11.20\Std{2.17} & 9.40\Std{1.97}
& 5.20\Std{1.17} & 7.80\Std{1.21} \\
explore\_obstructed\_goals\_small
& 2.52 & 10.00\Std{2.14} & 19.80\Std{2.29}
& 22.40\Std{2.78} & 26.80\Std{3.41}
& 12.80\Std{1.92} & 16.20\Std{2.10} \\
psychlab\_visual\_search
& 0.25 & 1.32\Std{0.16} & 3.82\Std{0.37}
& 8.72\Std{0.80} & 30.96\Std{1.75}
& 0.66\Std{0.12} & 8.24\Std{0.68} \\
rooms\_exploit\_deferred\_effects\_train
& 21.17 & 16.98\Std{2.75} & 15.14\Std{2.22}
& 13.20\Std{2.40} & 18.62\Std{3.04}
& 17.44\Std{2.99} & 12.48\Std{2.21} \\
rooms\_watermaze
& 14.48 & 9.34\Std{0.96} & 9.60\Std{1.52}
& 6.90\Std{1.07} & 14.84\Std{1.79}
& 1.80\Std{0.42} & 7.54\Std{0.95} \\
\midrule
Mean
& 3.79 & 7.53\Std{1.14} & 12.62\Std{1.48}
& 13.32\Std{1.55} & 18.86\Std{2.10}
& 8.94\Std{1.23} & 10.78\Std{1.38} \\
\bottomrule
\end{tabular}%
}
\end{table}

%%%%%%%%%%%%%%%%%%%%%%%%%%%%%%%%%%%%%%%%%%%%%%%%%%%%%%%%%%%%%%%%%%%%%%%%%%%%%%%

\section{Additional DMLab Ablations}
\label{app:dmlab_ablations}

This section reports DMLab ablations that complement the main analysis. We study
whether previous-action conditioning is responsible for the translation gains
and how decoded DMLab transitions change with retained prefix length.

\subsection{Translator Conditioning}
\label{app:translator_conditioning}

The translator conditions on latent tokens and the previous action. We ablate this
choice by comparing three inputs, namely latent tokens only, latent tokens plus previous
action, and latent tokens plus previous action and current observation. The same
conditioning choice is used for FlexLAM and fixed-capacity baselines in the main
comparisons.

\begin{figure}[ht]
  \centering
  \includegraphics[width=0.45\linewidth]{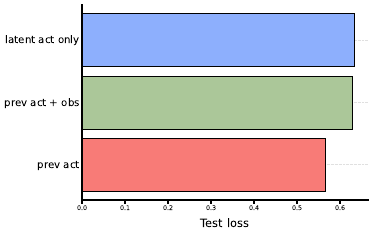}
  \caption{\textbf{Translator conditioning ablation.}
  Translator test loss for three input choices, namely $z$ only, $z$ + previous action, and
  $z$ + previous action + observation.
  Conditioning on $a_{t-1}$ improves prediction in egocentric settings. Directly
  feeding $o_t$ can make the translator more sensitive to appearance variation under
  limited supervision.}
  \label{fig:translator_conditioning_app}
\end{figure}

Routing through the latent transition representation reduces direct access to
task-specific appearance cues in $o_t$ and makes the conditioning path consistent
with the LAM interface used in the main experiments.

\subsection{DMLab Prefix-Length Reconstruction}
\label{app:dmlab_prefix_reconstruction}

We visualize how DMLab reconstructions change as the retained prefix length varies.
This diagnostic illustrates the prefix-valid structure induced by retained-prefix
training.

\begin{figure}[ht]
  \centering
  \includegraphics[width=\linewidth]{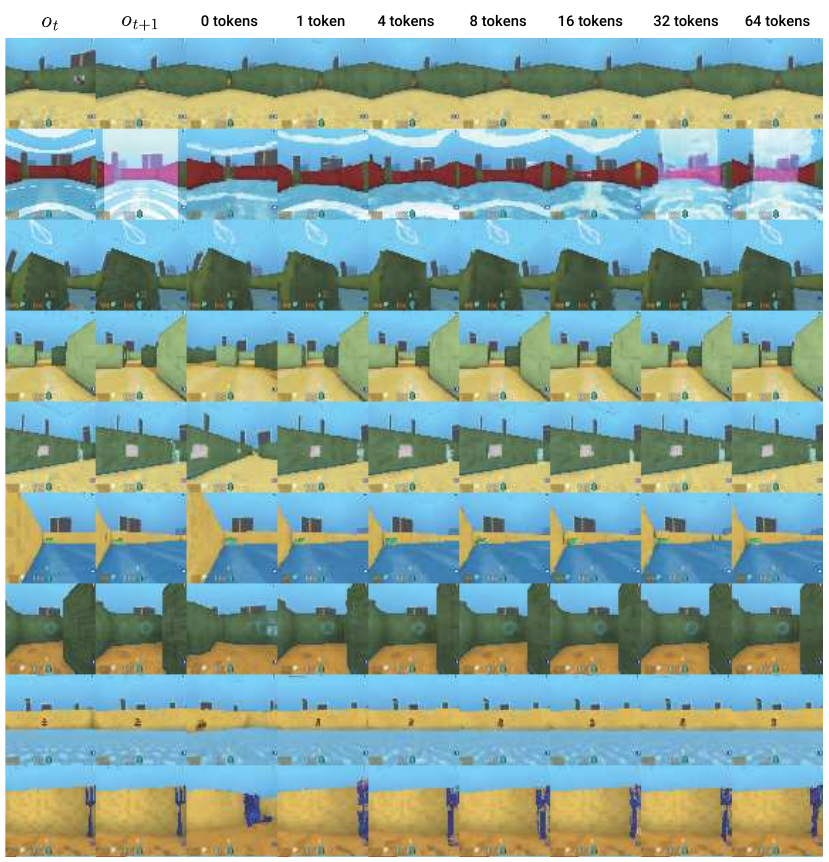}
  \caption{\textbf{DMLab prefix-length reconstruction.}
  Reconstruction results for the same DMLab transition while varying retained prefix
  length $k$. Increasing $k$ progressively recovers finer details, while small prefixes
  capture coarse transition structure.}
  \label{fig:dmlab_prefix_recon_app}
\end{figure}

%%%%%%%%%%%%%%%%%%%%%%%%%%%%%%%%%%%%%%%%%%%%%%%%%%%%%%%%%%%%%%%%%%%%%%%%%%%%%%%

\section{Real-World Video Pretraining and Evaluation Details}
\label{app:real_world_details}

\subsection{Data Mixture and Sampling}
\label{app:data_mixture_and_sampling}

We pretrain the real-world FlexLAM model on a mixture of Internet video, egocentric
video, and robot video. All videos are sampled at 1.6 fps, and adjacent sampled frames
are used as transition pairs $(o_t,o_{t+1})$. Robot video datasets are upsampled
relative to their raw size to ensure sufficient coverage of robot-video transitions.
The mixture is intentionally heterogeneous because it combines egocentric human
video, robot-video data, and Internet video. We use this setting to test whether the
retained-prefix bottleneck remains usable across diverse transition sources.

\begin{table}[ht]
\centering
\footnotesize
\setlength{\tabcolsep}{4pt}
\caption{\textbf{Pretraining data statistics.}
Data mixture for real-world and robot video pretraining. The weight column indicates
sampling ratios during training.}
\label{tab:pretrain_data_stats}
\begin{tabular}{@{}lrrrr@{}}
\toprule
Name & Rows & Weight & Orig. rows & Hours \\
\midrule
OXE~\citep{embodimentcollaboration2025openxembodimentroboticlearning}                & 1{,}570{,}254 & 3 &   523{,}418 & 3000h \\
AgiBot alpha~\citep{agibotworldcontributors2025agibotworldcolosseolargescale}
                                &   621{,}216   & 3 &   207{,}072 &  300h \\
AgiBot beta (40\%)~\citep{agibotworldcontributors2025agibotworldcolosseolargescale}
                                & 2{,}447{,}904 & 3 &   815{,}968 & 1200h \\
Ego4D~\citep{grauman2022ego4dworld3000hours}            & 4{,}464{,}867 & 1 & 4{,}464{,}867 & 3670h \\
OpenVidHD-0.4M~\citep{nan2025openvid1mlargescalehighqualitydataset} & 1{,}200{,}111 & 3 &   406{,}781 & 1200h \\
VIDGEN-1M~\citep{tan2024vidgen1mlargescaledatasettexttovideo}       & 4{,}012{,}652 & 4 & 1{,}003{,}163 & 2200h \\
HOIGen-1M~\citep{liu2025hoigen1mlargescaledatasethumanobject}       & 2{,}620{,}624 & 4 &   655{,}156 & 2200h \\
\bottomrule
\end{tabular}
\end{table}

\subsection{Model Initialization and Latent Injection}
\label{app:model_initialization}

For real-world video, the encoder is initialized with VideoMAE-v2 Large
~\citep{wang2023videomaev2scalingvideo}. The decoder is initialized with SD3~\citep{esser2024scalingrectifiedflowtransformers}.
We condition on the current observation $o_t$ through an image-conditioning pathway
and inject a null-filled retained-prefix representation into the conditioning pathway.
The bottleneck uses FSQ with maximum token length $K=80$.

\subsection{Real-World Decoder Objective Details}
\label{app:real_world_decoder_objective}

The real-world decoder uses the same retained-prefix conditioning principle as the
DMLab decoder. The decoder is conditioned on $o_t$ and
$\tilde{z}_t^{(k)}$, and is trained with a rectified-flow objective. In implementation,
the objective is a flow-matching velocity objective under a rectified-flow
formulation. Compared with DMLab, the real-world setting uses higher-resolution
inputs, pretrained initialization, and a larger bottleneck to accommodate greater
visual diversity.

\subsection{FlexLAM-Real Hyperparameters}
\label{app:flexlam_real_hparams}

Table~\ref{tab:flexlam_real_hparams} summarizes the real-world FlexLAM model.
The real-world setting differs from DMLab in resolution, initialization, and data
mixture, but uses the same retained-prefix bottleneck principle. The encoder is
initialized from VideoMAE-v2, the decoder from SD3, and the bottleneck uses
FSQ with maximum length $K=80$.

\begin{table}[ht]
\centering
\small
\caption{\textbf{FlexLAM-real hyperparameters.}
Initialization and training settings for the real-world video setting.}
\label{tab:flexlam_real_hparams}
\begin{tabularx}{\linewidth}{@{}llX@{}}
\toprule
Component & Parameter & Value \\
\midrule
Input & frame sampling & 1.6 fps \\
& image size & $224 \times 304$ \\
\midrule
Bottleneck & FSQ levels & [7, 5, 5, 5, 5] \\
& max tokens $K$ & 80 \\
& retained-prefix & training with $k \sim p(k)$ \\
\midrule
VAE & model & SD3.5 medium VAE \\
\midrule
Encoder & init & VideoMAE-v2 Large \\
& depth & 24 \\
& embed dim & 1024 \\
& mlp ratio & 4 \\
& tubelet size & 2 \\
\midrule
Decoder & init & SD3.5 medium \\
& num layers & 24 \\
& num attention heads & 24 \\
& attention head dim & 64 \\
& in\_channels & 32 (default $\times 2$) \\
& out\_channels & 16 \\
\midrule
Conditioning & $o_t$ injection & image-conditioning path \\
& null-filled retained-prefix representation & conditioning pathway \\
\midrule
Objective & decoder training & rectified flow \\
\midrule
Training & steps & 200k \\
& learning rate & $3\times 10^{-5}$ \\
& batch size & 1024 \\
& hardware & 8 NVIDIA H100 GPUs \\
& wall-clock time & approximately 350 hours \\
\bottomrule
\end{tabularx}
\end{table}

\subsection{Ego4D Evaluation Protocol}
\label{app:ego4d_protocol}

Table~\ref{tab:ego4d_eval_protocol} gives the protocol used for the Ego4D
transition-reconstruction comparison. The evaluation uses held-out clips and adjacent
sampled frames. The released villa-X-LAM checkpoint is evaluated at its native
fixed-bottleneck setting, while FlexLAM is evaluated at retained prefix lengths
$k\in\{5,20,80\}$.

\begin{table}[ht]
\centering
\small
\caption{\textbf{Ego4D reconstruction evaluation protocol.}}
\label{tab:ego4d_eval_protocol}
\begin{tabularx}{\linewidth}{@{}lX@{}}
\toprule
Item & Setting \\
\midrule
Dataset & Ego4D held-out clips~\citep{grauman2022ego4dworld3000hours} \\
Evaluated clips & 200 \\
Frame sampling & 1.6 fps \\
Input resolution & $224 \times 304$ \\
Evaluated transition & adjacent sampled frames \\
Metrics & PSNR, SSIM, LPIPS~\citep{zhang2018unreasonableeffectivenessdeepfeatures} \\
Averaging & per-frame average over evaluated clips \\
Prefix lengths & $k \in \{5,20,80\}$ \\
Baseline & released villa-X-LAM checkpoint~\citep{chen2025villaxenhancinglatentaction} \\
Baseline setting & native fixed bottleneck setting, 7 latent actions per 8-frame clip, VQ size 32 \\
\bottomrule
\end{tabularx}
\end{table}

\subsection{Real-World Nominal Bottleneck Capacity}
\label{app:real_world_capacity}

Table~\ref{tab:real_world_nominal_capacity} reports nominal discrete bottleneck
capacities for the released villa-X-LAM reference and FlexLAM-real. These values
document the bottleneck budgets used in the real-world evaluation.

\begin{table}[t]
\centering
\caption{\textbf{Nominal discrete bottleneck capacity in real-world evaluation.}
villa-X-LAM is reported at its native 8-frame-clip setting. FlexLAM-real uses FSQ
[7,5,5,5,5], with nominal vocabulary size 4375 per latent token. These values are
reported to document the nominal token budgets.}
\label{tab:real_world_nominal_capacity}
\begin{tabular}{lccc}
\toprule
Method / setting & Quantizer & Tokens & Nominal capacity \\
\midrule
villa-X-LAM, 8-frame clip & VQ size 32 & 7 & $7\log_2 32 = 35.0$ bits \\
FlexLAM-real $k=1$ & FSQ [7,5,5,5,5] & 5 & $5\log_2 4375 \approx 12.1$ bits \\
FlexLAM-real $k=5$ & FSQ [7,5,5,5,5] & 5 & $5\log_2 4375 \approx 60.4$ bits \\
FlexLAM-real $k=20$ & FSQ [7,5,5,5,5] & 20 & $20\log_2 4375 \approx 241.8$ bits \\
FlexLAM-real $k=80$ & FSQ [7,5,5,5,5] & 80 & $80\log_2 4375 \approx 967.2$ bits \\
\bottomrule
\end{tabular}
\end{table}

%%%%%%%%%%%%%%%%%%%%%%%%%%%%%%%%%%%%%%%%%%%%%%%%%%%%%%%%%%%%%%%%%%%%%%%%%%%%%%%

\section{Additional Real-World Visualizations}
\label{app:additional_real_world_visuals}

\subsection{Additional Real-World Prefix Sweeps}
\label{app:additional_real_world_prefix_sweeps}

We provide additional examples of real-world prefix sweeps. These visualizations
complement the Ego4D metrics in Table~\ref{tab:ego4d_metrics} by showing how
decoded transitions change as more latent-action tokens are retained. The examples
include egocentric video and robot-video clips, so they illustrate the same
prefix-valid behavior under camera motion, background variation, and object
interaction.

\begin{figure}[ht]
  \centering
  \includegraphics[width=\linewidth]{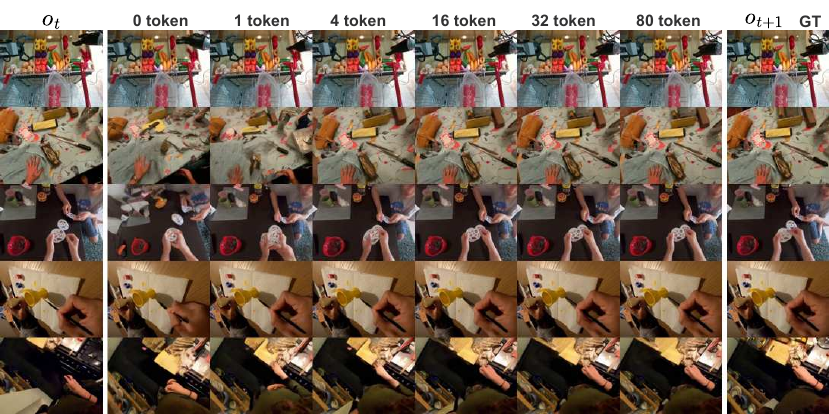}
  \caption{\textbf{Additional real-world prefix sweeps.}
  Reconstructions from the same FlexLAM model while varying retained prefix length
  $k$ across Ego4D and robot-video examples. Larger prefixes recover additional visual
  detail, while shorter prefixes preserve coarse transition structure.}
  \label{fig:additional_real_world_prefix_sweeps}
\end{figure}

%%%%%%%%%%%%%%%%%%%%%%%%%%%%%%%%%%%%%%%%%%%%%%%%%%%%%%%%%%%%%%%%%%%%%%%%%%%%%%%

\section{Extended Impact Details}
\label{app:limitations_impact}

Large-scale video pretraining may involve private, copyrighted, biased, or
geographically imbalanced content. Dataset curation, filtering, licensing, consent,
and distribution-shift evaluation are important before applying similar pipelines.
Decoded frames should not be treated as factual evidence of real events.

\newpage

\end{document}